\documentclass[10pt,twocolumn,letterpaper]{article}

\usepackage[pagenumbers]{wacv} 
\usepackage[accsupp]{axessibility}  

\usepackage{graphicx}
\usepackage{amsmath}
\usepackage{amssymb}
\usepackage{booktabs}

\usepackage{amsfonts}
\usepackage{setspace}
\usepackage{enumitem}
\usepackage{setspace}
\usepackage{amssymb}
\usepackage{epsfig}
\usepackage{graphicx}
\usepackage{xcolor}
\usepackage{algorithm}
\usepackage{algorithmicx}
\usepackage{algpseudocode}
\usepackage{amsmath}
\usepackage{cite}

\floatname{algorithm}{Algorithm}
\renewcommand{\algorithmicrequire}{\textbf{Offline:}}
\renewcommand{\algorithmicensure}{\textbf{Online:}}

\usepackage[ps2pdf,bookmarks=false,
bookmarksnumbered=false, 
bookmarksopen=false,
colorlinks=false]{}

\usepackage[utf8]{inputenc}
\usepackage[english]{babel}
\usepackage{amsthm}
\theoremstyle{definition}

\usepackage{tikz}
\newsavebox{\tempbox}

\usepackage{multicol}

\usepackage[pagebackref,breaklinks,colorlinks]{hyperref}

\usepackage[capitalize]{cleveref}
\crefname{section}{Sec.}{Secs.}
\Crefname{section}{Section}{Sections}
\Crefname{table}{Table}{Tables}
\crefname{table}{Tab.}{Tabs.}

\def\wacvPaperID{980} 
\def\confName{WACV}
\def\confYear{2024}

\begin{document}

\title{\vspace{-1.2cm} Maximum Knowledge Orthogonality Reconstruction \\ with Gradients in Federated Learning \vspace{-0.65cm}}

\author{Feng Wang, Senem Velipasalar, and M. Cenk Gursoy \\
        EECS department, Syracuse University, Syracuse, NY, 13244.\\
        {\tt\small \{fwang26, svelipas, mcgursoy\}@syr.edu}
        }
        \vspace{-0.76cm}
\maketitle

\begin{abstract}
\vspace{-0.1cm}
Federated learning (FL) aims at keeping client data local to preserve privacy.
Instead of gathering the data itself, the server only collects aggregated gradient updates from clients. 
Following the popularity of FL, there has been considerable amount of work revealing the vulnerability of FL approaches by reconstructing the input data from gradient updates.~Yet, most existing works assume an FL setting with unrealistically small batch size, and have poor image quality when the batch size is large. Other works modify the neural network architectures or parameters to the point of being suspicious, and thus, can be detected by 
clients. Moreover, most of them can only reconstruct one sample input from a large batch. 
To address these limitations,
we propose a novel and analytical approach, referred to as the maximum knowledge orthogonality reconstruction (MKOR), to reconstruct clients' data. Our proposed method reconstructs a mathematically proven high-quality image from large batches. MKOR only requires the server to send secretly modified parameters to clients and can efficiently and inconspicuously reconstruct images from clients' gradient updates.
We evaluate MKOR's performance on MNIST, CIFAR-100, and ImageNet datasets and compare it with the state-of-the-art baselines. The results show that MKOR outperforms the existing approaches, and draw attention to a pressing need for further research on the privacy protection of FL so that comprehensive defense approaches can be developed. Code implementation available at: https://github.com/wfwf10/MKOR.
\end{abstract}

%
%
%
\vspace{-0.8cm}
\section{Introduction}  \label{sec:intro}
\vspace{-0.05cm}
Federated learning (FL), as a type of distributed learning, has attracted growing interest from both academia and industry. In the FL framework, multiple clients transmit gradient updates, instead of their data, to preserve client's privacy. In a typical setting, a parameter server broadcasts current network parameters to a subset of individual clients. Each selected client generates the gradient update based on its own local data batch and uploads it back to the server. Then, the server aggregates the gradient updates from selected clients and iterates until convergence~\cite{konevcny2016federated, bonawitz2017practical, kairouz2021advances}. 

However, it has been shown that FL is vulnerable to adversarial privacy attacks~\cite{geiping2020inverting, zhu2019deep}. By receiving gradient updates, an adversarial server can gain information about or reconstruct the private input data.
Some threat models consider an ``honest-but-curious" server that trains the model normally while recovering input data from unrealistically small batches~\cite{geiping2020inverting, zhu2019deep}. Other approaches consider a malicious server that modifies the network structure or parameters to deal with larger batches \cite{fowl2021robbing, wen2022fishing}, but these suspicious modifications might be detected and rejected by the clients. 

Motivated by the aforementioned limitations of existing works, we propose a novel and analytical approach, referred to as the maximum knowledge orthogonality reconstruction (MKOR), to reconstruct better quality input data with larger batches.~During common FL training, clients upload batches of gradients to a parameter server, to cooperatively train a neural network~\cite{kairouz2021advances}.~We consider a malicious server that uses the original structure of a convolutional neural network (CNN), but modifies the parameters instead
to maximize the orthogonality between prior- and post-knowledge.~The general assumption is that neighboring pixels have similar intensity values, which we define as the 
prior-knowledge.~By receiving gradient updates for broadcasted network parameters, a malicious server obtains further information about the unknown data, which we define as the post-knowledge.~The server optimizes reconstruction performance for multiple inputs from large batches.~We propose a malicious parameter design on the fully-connected (FC) classifier to decouple a small subset of nodes, which broadcasts analytic attacks from a single FC layer to multiple FC layers.~Then, we define the knowledge orthogonality to judge the efficiency of malicious parameters in convolutional layers, and demonstrate the corresponding parameter design on the convolutional layers in a VGG model~\cite{simonyan2014very} to optimize reconstruction.~We also demonstrate how these modifications can be performed inconspicuously to avoid being detected by clients, and show that such modifications can easily apply to other CNNs with similar architectures, such as LeNet~\cite{lecun1998gradient}.~We evaluate the performance of MKOR on MNIST dataset~\cite{deng2012mnist} with LeNet5 network, and CIFAR-100~\cite{krizhevsky2009learning} and ImageNet~\cite{deng2009imagenet} datasets with the VGG16 network, then compare it with existing works. 

Main contributions of this work include the following: \vspace{-0.18cm}
\begin{itemize}[leftmargin=0.3cm,
itemsep=0.1pt
]
    \item In contrast to most optimization-based attacks, MKOR can recover private data from arbitrarily large batch sizes.\vspace{-0.15cm} 
    \item Unlike optimization- and semi-optimization-based attacks, MKOR is completely analytical, and provides guaranteed high-fidelity reconstruction from arbitrary data distribution.~We calculate the mathematical solution without iterative optimization or generating gradient updates.\vspace{-0.15cm}
    \item In contrast to malicious threat models, which suspiciously modifies a network's structure, MKOR uses the original architecture, which is VGG16 or LeNet in this paper.\vspace{-0.15cm}
    \item Compared to malicious threat models, which only modify the parameters, (i) MKOR is inconspicuous and hard to detect, since there is not a large number of zero weights dominating any layer, and the sum of gradients is not dominated by any sample (there is no weight changing within one batch for different samples either); and (ii) MKOR is more efficient, since it generates multiple reconstructions with one gradient update. \vspace{-0.1cm}
\end{itemize}
%
%
%
\section{Related Work} \label{sec:related}
\vspace{-0.05cm}
\begin{table*}[t]
    \small
	\begin{center}
	    \caption{Comparison between different methods of input reconstruction attack}
     \vspace{-0.3cm}
     \resizebox{0.84\textwidth}{!}{%
		\begin{tabular}{ | l |  l | l | l | l |  l | } \hline
			Methods & Type & Assumption on server & \vtop{\hbox{\strut Maximum}\hbox{\strut batch size}} & \vtop{\hbox{\strut Samples recovered}\hbox{\strut per batch}}  & \vtop{\hbox{\strut Generalization to}\hbox{\strut unknown datasets}}   \\   \hline
            Gradient inversion \cite{geiping2020inverting} & Optimization & Honest-but-curious & $\leq 8$ & $\leq 8$ & Trained on client data  \\   \hline
            Deep leakage \cite{zhu2019deep} & Optimization & Honest-but-curious & $\leq 8$ & $\leq 8$ & Trained on client data  \\   \hline
            Other earlier works~\cite{wang2019beyond,zhao2020idlg, yin2021see} & Optimization & Honest-but-curious & $\leq 8$ & $\leq 8$ & Restricted  \\   \hline
            Modify network structure \cite{fowl2021robbing, boenisch2021curious} & Analytical & Malicious-very easy to detect & $> 100$ & varies & Data-free (no training)   \\   \hline
            Class fishing \cite{wen2022fishing} & Optimization & Malicious-easy to detect & $> 100$ & $\leq 1$ & Restricted   \\   \hline
            MKOR (Ours) & Analytical & Malicious-and-inconspicuous & $> 100$ & $\leq 100$ & Data-free (no training)\\   \hline
		\end{tabular}
  }
\label{tab:other_methods}
\end{center}
\vspace{-0.8cm}
\end{table*}

We consider a common FL framework, where a parameter server orchestrates a set of $U$ clients to cooperatively train a neural network~\cite{kairouz2021advances}.~Each client $u$ has a local dataset with $K_u$ images, where the $k^{th}$ image $\textbf{x}_{u,k}$ is assigned the label $y_{u,k}$.~The goal of FL is to iteratively optimize the neural network parameters $\textbf{A}$ (the weights and biases) to minimize the sum of loss $\mathcal{L}$ via $\operatorname*{argmin}_{\textbf{A}} \sum_{u}^{U} \sum_{k}^{K_u} \mathcal{L}(\textbf{x}_{u,k}, y_{u,k} \, | \, \textbf{A})$, 
where each client $u$ uploads the local sum of gradients $\sum_{k}^{K_u} \nabla_{\textbf{A}} \mathcal{L}(\textbf{x}_{u,k}, y_{u,k} \, | \, \textbf{A})$, instead of original data $\cup_k^{K_u} \{\textbf{x}_{u,k}, y_{u,k}\}$, to the server. In the remainder of this paper, we consider the data $\{\textbf{x}_k, y_k\}$ and the gradients of each single client batch with size $K$, and omit the client index $u$. The list of notations is provided in Tab.~1 in the Suppl. file.

There have been different works~\cite{melis2019exploiting, wainakh2022user,zhu2019deep,geiping2020inverting} suspecting the central server to be curious or malicious with the goal of obtaining clients' data. 
While some methods~\cite{melis2019exploiting, wainakh2022user} investigate reconstructing the labels $y_{k}$ from gradients, many studies demonstrate a privacy attack
that reconstructs the client input $\textbf{x}_{k}$ from the gradients, which is usually referred to as the ``gradient inversion" \cite{geiping2020inverting} or ``deep leakage"\cite{zhu2019deep}.
Most of these works consider the threat of an ``honest-but-curious" server, which not only maintains the FL functionality and updates the parameters as planned, but also intends to reconstruct client inputs from the received gradients.~After an optimization-based attack was proposed in~\cite{wang2019beyond}, majority of the ensuing attacks were designed in the same fashion~\cite{zhu2019deep, geiping2020inverting, zhao2020idlg, yin2021see}.~Given the well-trained parameters $\textbf{A}$, these attacks randomly initialize dummy data $\textbf{x}'_k$ and labels $y'_k$, generate dummy gradient $\sum_{k}^K \nabla_{\textbf{A}} \mathcal{L}(\textbf{x}'_k, y'_k \, | \, \textbf{A})$, and iteratively update the dummy data with its dummy gradient.
%

Yet, the aforementioned approaches can only reconstruct from gradients of batches with very limited batch size. This restricts their effectiveness and usability in realistic settings, where the gradient is aggregated over many samples~\cite{kairouz2021advances}.~There are other
assumptions undermining their practicality.~For instance, they are less likely to succeed in the absence of FL convergence, and some of them make a strong assumption about knowing the Batch-Norm statistics and private labels~\cite{huang2021evaluating}.~It should also be noted that optimization-based methods have no guarantee of convergence.~While the best performing sample over a dataset can be recognizable, the performance on other samples can be disappointing. Moreover, some other works~\cite{yin2021see, hatamizadeh2022gradvit}, including generative adversarial network (GAN)-based attacks \cite{li2022auditing}, use a pre-trained network to reconstruct high-resolution images in relatively small batches where the batch sizes are far less than the number of classes. Most of these methods have strong assumption on the correlation between the distributions of the attacker's training dataset and the user dataset, which is barely true in FL. In contrast, with  MKOR, the attacker does not need surrogate data. 

Other threat models consider a malicious server that can manipulate the network structure or  parameters.
Yet,  
these modifications 
can be deemed suspicious and easily detectable. Privacy attacks in~\cite{fowl2021robbing, boenisch2021curious} construct large FC layers whose output size is similar to the input image, and reconstruct inputs with a large batch size. 
Yet, such a large FC layer, which is larger than a typical CNN, is highly suspicious and thus can be rejected, since the purpose of using conv. layers is to reduce the FC layer size. Attacks in \cite{pasquini2022eluding} and the feature fishing attack in \cite{wen2022fishing} even require the server to send different malicious parameters to the same client, to be applied on different samples in the same batch.
Such modification is too suspicious and easily detectable. The class fishing attack in \cite{wen2022fishing} keeps the network structure intact, and modifies the weights in the last FC layer by setting all but those for the target label to zero. Hence, the sum gradient is dominated by one sample and can be easily detected. For example on ImageNet, it sets 99.9\% of weights to zero and 99.9\% of images have zero gradient, while our MKOR only modifies at most 14\% of weights to non-zero values in a layer 
and every image has similar (non-zero) gradient magnitude.  
The attack in \cite{lam2021gradient} requires the index of each sample as input (which is not necessary for FL). It also requires excessively large number of iterations on the same user with the same training batch, while most clients might only appear once in cross-device FL of that batch size \cite{kairouz2021advances}. 

It should be noted that, against privacy attacks, every client should verify the received network before deploying their data. Thus, when an anomaly is detected, clients may exit the FL process before uploading the gradients.~The detection methods include checking the size of the FC layer,
checking if the majority of weights are $0$ in each layer, and sampling the gradients 
to see if the sum gradient is dominated by a few of them. These detection processes are 
easy to implement, and can detect all the aforementioned attacks. We also note that data distribution is mostly hidden from the server, and most of the optimization-based methods depend on the well-trained parameters with the clients' data.

In this work, we also focus on the reconstruction of the input $\textbf{x}_{k}$ from the gradients, and demonstrate that MKOR can inconspicuously achieve guaranteed reconstruction performance on large batches with original/unmodified network models without any assumptions on data distribution. A general comparison of our approach with the aforementioned methods is provided in Tab.~\ref{tab:other_methods}, illustrating the advantages of MKOR addressing the prior limitations. 
%
%
%
\vspace{-0.15cm}
\section{Maximum Knowledge Orthogonality Reconstruction} \label{sec:MKOR}
\vspace{-0.1cm}
We introduce MKOR as a novel and fully analytical method to reconstruct inputs $\textbf{x}_k$ from the sum gradient $\sum_{k}^K \textbf{G}_k$, where $\textbf{G}_k = \nabla_{\textbf{A}} \mathcal{L}(\textbf{x}_k, y_k \, | \, \textbf{A})$.~We consider a malicious server that modifies the parameters $\textbf{A}$ 
of a CNN and can handle large batches.~The proposed MKOR is inconspicuous and avoids being detected by clients.~We explain details by using the original VGG16 architecture \cite{simonyan2014very}, and then apply it on other networks such as LeNet.
\vspace{-0.1cm}
\subsection{Sum Gradient Knowledge Shrinkage} \label{subsec:necessary}
\vspace{-0.05cm}
We first show the necessity to decouple gradients. Given a certain distribution of the gradient vector $\textbf{G}$, we can denote the information leakage about $\textbf{G}_1$, caused by revealing the sum gradient, as $H(\textbf{G}_1) - H(\textbf{G}_1 | \sum_{k=1}^{K} \textbf{G}_k)$, where $H(\cdot)$ is the entropy. Compared to a scenario with a small batch size, where leakage is significant, it is obvious that as batch size $K\to\infty$, 
%
%
$\lim\limits_{K \to\infty} \frac{1}{K} \sum_{k=1}^{K} \textbf{G}_k = \lim\limits_{K \to\infty} \frac{1}{K-1} \sum_{k=2}^{K} \textbf{G}_k = \mathbb{E}(\textbf{G})$,
where $\mathbb{E}(\cdot)$ denotes the expectation. Therefore, 
\vspace{-0.25cm}
\begin{equation} \label{eq:inf_leakage_converge}
\footnotesize{
\lim\limits_{K \to\infty} \left( H(\textbf{G}_1) - H(\textbf{G}_1 | \sum_{k=1}^{K} \textbf{G}_k) \right) = 0,
}
\vspace{-0.15cm}
\end{equation}
and the knowledge leakage from batch-normalized average gradient goes to 0 (except the parameters in the last FC layer and the decoupled parameter $n$ in $l^{th}$ layer, where $\exists k, \forall k' \neq k, \textbf{G}_{k'}^l[n] = 0$. Decoupling all the parameters can be detected by clients easily and is not recommended. This will be further explained in Sec.~\ref{subsec:dense}). 

Specifically, if we assume that each element $\textbf{g} \in \textbf{G}$ follows a Gaussian distribution $\mathcal{N}(\mu_g,\,\sigma_g^{2})$, then each element in sum gradient $\sum_{k=1}^{K} \textbf{g}_k \sim \mathcal{N}(K \mu_g,\, K\sigma_g^{2})$, and $H(\sum_{k=1}^{K} \textbf{g}_k)=\frac{1}{2}\log(2\pi K \sigma_g^2)+\frac{1}{2}$. Subsequently, for $K>1$, we have \vspace{-0.3cm}
\begin{equation} \label{eq:finite_leakage_converge}
\scriptsize{
H(\textbf{g}_1) - H(\textbf{g}_1 | \sum_{k=1}^{K} \textbf{g}_k) = H(\sum_{k=1}^{K} \textbf{g}_k) - H(\sum_{k=1}^{K} \textbf{g}_k | \textbf{g}_1) =  O\left( \frac{1}{K} \right)
}
\vspace{-0.25cm}
\end{equation}
where $O(\frac{1}{K})$ denotes $K^{-1}$ complexity. The derivation is provided in the Suppl. material.   

Thus, when the batch size is large, it is necessary to decouple the parameters
so that a subset of the gradients originates from samples with the same label. We next introduce our design modifying the FC classifier parameters to achieve the best reconstruction performance.
\vspace{-0.05cm}
\subsection{FC Classifier Input Reconstruction} \label{subsec:dense}
\vspace{-0.1cm}
First, we note that a linear combination of an FC classifier input from each class can be reconstructed. Consider a ``vanilla" feed-forward FC classifier, which can be regarded as a first-order Markov process. By reconstructing the input of the FC classifier, we will have all the information we need to reconstruct the input of the convolutional part. 

Also, there can be ample amount of samples such that they are the only ones representing their class in the batch, and thus, the aforementioned reconstruction becomes not just a linear combination, but can uniquely recover the sample information. Assume that the local sample distribution over classes of a client is $\textbf{p} = \{p_1, p_2, \ldots, p_{N}\}$, where $N$ is the number of labels. The expectation of the number of samples that are the only ones representing their class is  \vspace{-0.3cm}
\begin{multline} \label{eq:exp_single_sample}
\scriptsize{
\begin{aligned}
& \sum_{n=1}^N \sum_{k=1}^K \left( \text{P}(y_k=n) \prod_{k' \neq k}^K \text{P}(y_{k'} \neq n) \right) \\
= & \sum_{n=1}^N \sum_{k=1}^K p_n (1 - p_n)^{K-1} = K \sum_{n=1}^N p_n (1 - p_n)^{K-1} .  \\
\end{aligned}
\vspace{-0.4cm}
}
\vspace{-0.3cm}
\end{multline}
For example,~if there are $K\!\!=\!\!100$ samples for $N\!\!\!=\!\!\!100$ classes with a uniform distribution, we expect 36.97 unique samples. Thus, we can reconstruct a significant amount of samples that are not combined with others.
We now explain how the input is reconstructed without error. Let $L$ denote the number of layers and each layer $l$ has $N^l$ input nodes (most superscripts indicate the layer index as listed in Tab.~1 in the Suppl.~file). The weights and biases are denoted by $\textbf{W}^l \in \mathbb{R}^{N^{l+1} \times N^l}$ and $\textbf{b}^l \in \mathbb{R}^{N^{l+1}}$, where the parameters $\textbf{A}^l = \{\textbf{W}^l, \textbf{b}^l\}$ and $N_{L+1} \stackrel{\text{def}}{=} N$. For each layer $l$, we denote the input of the FC classifier from sample $k$ as $\textbf{z}_k^l \in \mathbb{R}^{N^l}$, and thus the output of conv. layers is $\textbf{z}_k^1 \in \mathbb{R}^{N_1}$.
\vspace{-0.4cm}
\subsubsection{Single Layer} \label{subsubsec:dense_single}
\vspace{-0.2cm}
If the FC classifier consists of only one layer without an activation function, it is easy to recover the input without modifying the parameters~\cite{geiping2020inverting, zhu2020r}. For the $k^{th}$ sample with output vector $\textbf{y}_k$ with assigned label $y_k$,
\begin{equation} \label{eq:1_layer_sol_1}
\footnotesize{
\frac{\partial \mathcal{L}(\textbf{x}_k, y_k \, | \, \textbf{A})}{\partial \textbf{W}^1} = \frac{\partial \mathcal{L}(\textbf{x}_k, y_k \, | \, \textbf{A})}{\partial \textbf{y}_k} \frac{\partial \textbf{y}_k}{\partial \textbf{W}^1} = \frac{\partial \mathcal{L}(\textbf{x}_k, y_k \, | \, \textbf{A})}{\partial \textbf{b}^1} {\textbf{z}_k^1}^\text{T}, 
}
\vspace{-0.1cm}
\end{equation}
where $\frac{\partial \mathcal{L}(\textbf{x}_k, y_k \, | \, \textbf{A})}{\partial \textbf{y}_k} = \frac{\partial \mathcal{L}(\textbf{x}_k, y_k \, | \, \textbf{A})}{\partial \textbf{b}^1}$ since $\textbf{y}_k = \textbf{W}^1 \textbf{z}_k^1 + \textbf{b}^1$. Hence, we have 
\begin{equation} \label{eq:1_layer_sol_2}
\footnotesize{
\textbf{z}_k^1 = \frac{\partial \mathcal{L}(\textbf{x}_k, y_k \, | \, \textbf{A})}{\partial \textbf{W}^1[y_k, :]} / \frac{\partial \mathcal{L}(\textbf{x}_k, y_k \, | \, \textbf{A})}{\partial \textbf{b}^1[y_k]} ,
\vspace{-0.15cm}
}
\end{equation}
where $\textbf{W}^1[y_k, :]$ denotes the $y_k^{th}$ row of $\textbf{W}^1$, and the division is element-wise. In practice, an attacker may obtain $\sum_{k=1}^K \frac{\partial \mathcal{L}(\textbf{x}_k, y_k \, | \, \textbf{A})}{\partial \textbf{A}}$ instead of that for a single sample, so the attacker will reconstruct
\begin{equation} \label{eq:1_layer_sol_3}
\vspace{-0.1cm}
\footnotesize{
\hat{\textbf{z}}_n^1 = \sum_{k=1}^{K} \frac{\partial \mathcal{L}(\textbf{x}_k, y_k \, | \, \textbf{A})}{\partial \textbf{W}^1[y_k, :]} / \sum_{k=1}^{K} \frac{\partial \mathcal{L}(\textbf{x}_k, y_k \, | \, \textbf{A})}{\partial \textbf{b}^1[y_k]},
\vspace{-0.15cm}
}
\end{equation}
where $\hat{\textbf{z}}_n^1 = \textbf{z}_k^1 \neq \textbf{0}$ if and only if there is exactly one $k$ such that $y_k=n$. Thus, the input of a linear layer can be accurately reconstructed. The reason this reconstruction method works is that there exists one \textbf{decoupled} element in the output vector $\textbf{y}_k$ for each label $n$, i.e., 
\vspace{-0.1cm}
\begin{equation} \label{eq:1_layer_decouple}
\footnotesize{
\sum\limits_{\substack{k=1 \\ y_k = n}}^{K} \frac{\partial \mathcal{L}(\textbf{x}_k, y_k \, | \, \textbf{A})}{\partial \textbf{y}_k[n]} \neq \textbf{0} , \quad
\sum\limits_{\substack{k=1 \\ y_k \neq n}}^{K} \frac{\partial \mathcal{L}(\textbf{x}_k, y_k \, | \, \textbf{A})}{\partial \textbf{y}_k[n]} = \textbf{0} .
}
\vspace{-0.2cm}
\end{equation}
\subsubsection{Multiple Layers - Naive Decoupling}\label{subsubsec:dense_intuitive}
\vspace{-0.2cm}
\begin{figure}
	\centering
	\includegraphics[width=0.75\linewidth]{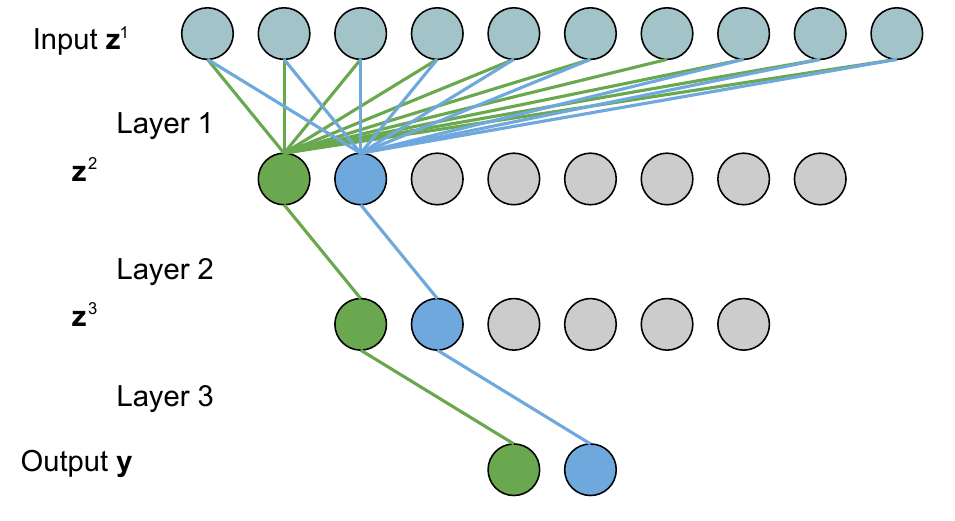}
 \vspace{-0.25cm}
	\caption{\small{A naive approach to decouple the second layer of multi-layer FC classifier.}}
	\label{fig:dense_decouple}
 \vspace{-0.4cm}
\end{figure}
While gradients in a single-layer classifier are easily decoupled, parameter modification is required to decouple the output of the first layer for a multi-layer classifier. A naive and also suspicious approach to decouple the output $\textbf{z}_k^2$ of the first layer in an FC classifier with multiple layers is illustrated in Fig. \ref{fig:dense_decouple}. If all weights, except the connected ones, are set to zero and the back-propagation is not zeroed out by the activation function, we have the following for each label $n \in \{1, 2, \ldots, N\}$: \vspace{-0.15cm}
\begin{equation} \label{eq:multi_layer_decouple_intuitive}
\vspace{-0.1cm}
\footnotesize{
\sum\limits_{\substack{k=1 \\ y_k = n}}^{K} \frac{\partial \mathcal{L}(\textbf{x}_k, y_k \, | \, \textbf{A})}{\partial \textbf{z}^2_k[n]} \neq \textbf{0} , \quad
\sum\limits_{\substack{k=1 \\ y_k \neq n}}^{K} \frac{\partial \mathcal{L}(\textbf{x}_k, y_k \, | \, \textbf{A})}{\partial \textbf{z}^2_k[n]} = \textbf{0} .
}
\vspace{-0.05cm}
\end{equation}
Similar to Eq.~(\ref{eq:1_layer_sol_1})(\ref{eq:1_layer_sol_2})(\ref{eq:1_layer_sol_3}), we have 
\vspace{-0.2cm}
\begin{equation} \label{eq:multi_layer_sol_intuitive}
\footnotesize{
\hat{\textbf{z}}_n^1 = \sum_{k=1}^{K} \frac{\partial \mathcal{L}(\textbf{x}_k, y_k \, | \, \textbf{A})}{\partial \textbf{W}^1[y_k, :]} / \sum_{k=1}^{K} \frac{\partial \mathcal{L}(\textbf{x}_k, y_k \, | \, \textbf{A})}{\partial \textbf{b}^1[y_k]} .
}
\vspace{-0.1cm}
\end{equation}
While this naive approach decouples $N$ elements in $\textbf{z}_k^2$, and thus, can reconstruct the input $\hat{\textbf{z}}_n^1$, the modification on the parameters with the given pattern and massive number of zeros are too suspicious and can easily be detected by clients. Hence, we propose an inconspicuous decoupling approach next, which achieves the same result. 
\vspace{-0.3cm}
\subsubsection{Multiple Layers - Inconspicuous Decoupling} \label{subsubsec:dense_inconspicuous}
\vspace{-0.15cm}
Consider a multi-layer FC classifier with ReLU activation. Our goal is to decouple a set of nodes $\{n_1, n_2, \ldots, n_N\}$ in $\textbf{z}_k^2$, where $\textbf{z}_k^2[n_i]$ uniquely corresponds to one output label as described in (\ref{eq:multi_layer_decouple_intuitive}). While all indices should be picked randomly, we list them in order (i.e., $n_1 \leq 2 < n_2 \leq 4 < \ldots < n_N \leq 2N$, similar to Fig. \ref{fig:dense_decouple}) for easier understanding. The parameters in the first layer are modified to
\vspace{-0.3cm}
\begin{align}\label{eq:1st_layer_weight}
\small{
\textbf{W}^1[n, :] = 
\begin{cases}
\alpha_n \textbf{W}^1[n \! - \! 1, :],\, \text{if } n \! \leq \! 2N, n \text{ mod } 2 = 0 \\
\textbf{W}^1[n, :],\, \text{otherwise} 
\vspace{-0.2cm}
\end{cases}
}
\end{align}
\vspace{-0.4cm}
\begin{align}\label{eq:1st_layer_bias}
\small{
\textbf{b}^1[n] = 
\begin{cases}
\alpha_n \textbf{b}^1[n - 1],\, \text{if } n \leq 2N, n \text{ mod } 2 = 0 \\
\textbf{b}^1[n],\, \text{otherwise}
\vspace{-0.2cm}
\end{cases} 
}
\end{align}
where $\alpha_n<0$ is a constant, $N$ is the number of labels, and $n \text{ mod } 2 = 0$ denotes $n$ to be an even number. Thus, it is guaranteed to have $N$ different indices $n \leq 2N$, where each of them indexes a non-zero node $\textbf{z}_k^2[n]$ given any unknown sample $k$. For each following layer $l>1$, the parameters are modified to
\vspace{-0.3cm}
\begin{align}\label{eq:latter_layer_weight}
\scriptsize{
\textbf{W}^l[n^{l+1}, n^l] = 
\begin{cases}
\left| \textbf{W}^l[n^{l+1}, n^l] \right| ,\, \text{if } l=2, \lfloor \frac{n^l}{2} \rfloor = n^{l+1} \leq N \\
\mathcal{N}(0,\,\sigma^{2}) ,\, \text{if } l=2, \lfloor \frac{n^l}{2} \rfloor \neq n^{l+1}, n^l \leq 2N \\
\left| \textbf{W}^l[n^{l+1}, n^l] \right| ,\, \text{if } l>2, n^l = n^{l+1} \leq N \\
\mathcal{N}(0,\,\sigma^{2}) ,\, \text{if } l>2, n^l \neq n^{l+1}, n^l \leq N \\
\textbf{W}^l[n^{l+1}, n^l],\, \text{otherwise} 
\vspace{-0.1cm}
\end{cases} 
}
\end{align}
\vspace{-0.8cm}
\begin{align}\label{eq:latter_layer_bias}
\footnotesize{
\textbf{b}^1[n^{l+1}] = 
\begin{cases}
\left| \textbf{b}^1[n^{l+1}] \right| ,\, \text{if } n^{l+1} \leq N \\
\textbf{b}^1[n^{l+1}],\, \text{otherwise}
\vspace{-0.2cm}
\end{cases} 
}
\end{align}
where $|\cdot|$ denotes the absolute value, $\lfloor \cdot \rfloor$ denotes rounding down, and $\mathcal{N}(0,\,\sigma^{2})$ denotes independent and identically distributed (i.i.d.) Gaussian noise with zero mean and variance $\sigma^{2} \ll 1$. Eq. (\ref{eq:latter_layer_weight}) ensures that first $2N$ nodes in the 2nd layer and first $N$ nodes in the following layers are decoupled to avoid any desired node being zeroed out by ReLU activation. Thus, each output element $\textbf{y}_k[n]$ has a unique back-propagation path to $\textbf{z}_k^2[2n-1]$ and $\textbf{z}_k^2[2n]$, and one of them is guaranteed to be decoupled for any sample. Since the majority of the parameters in (\ref{eq:latter_layer_weight}) and (\ref{eq:latter_layer_bias}) are not modified and the statistics of all parameters, such as mean and variance, are almost unchanged, it is extremely hard to detect the modification via parameters. Also, we note that such decoupling only affects a small subset of parameters and is hard to detect via gradient values, since the ReLU activation naturally eliminates gradients from many nodes. For better defense against any potential pattern-based detection, we may add a small i.i.d. noise to every parameter.

With the reconstructed input features $\hat{\textbf{z}}_n^1$, it is possible to train an auto-encoder, use the encoder as the convolutional part for a CNN, and use the decoder to reconstruct the input image $\hat{\textbf{x}}_{n}$.~Yet, the performance highly depends on the similarity of the 
local training set and the client dataset distributions. Instead, we introduce the analytical malicious parameter approach without any assumptions on data distribution. Before explaining how we set the malicious parameters of the convolutional layers and reconstruct the image, we first introduce the concept of knowledge orthogonality. 
\vspace{-0.1cm}
\subsection{Knowledge Orthogonality}\label{subsec:orthogonal}
\vspace{-0.15cm}
If gradients from none of the samples are zeroed out (otherwise the server will be suspicious), the gradient update of conv layers cannot be decoupled, and will be a summation over all samples. In Eq.~(\ref{eq:inf_leakage_converge})(\ref{eq:finite_leakage_converge}), we proved that the knowledge from such sum of gradients decreases and converges to zero as the batch size increases, which is the reason that most existing approaches do not perform well with large batches. In our approach, we alternatively use the extracted features $\hat{\textbf{z}}_n^1$ from conv layers with malicious parameters to reconstruct the input $\hat{\textbf{x}}_{n}$ for each label $n$, and introduce the knowledge orthogonality to judge the malicious parameters.

While there are edges and noise, images are mostly smooth, i.e., it is highly probable that neighboring pixels have similar intensity values~\cite{thum1984measurement, sayood2002lossless, lalush2004iterative}.~Also, the distribution of the intensities $x$ and $x'$ of two neighboring pixels from a set of images is likely to have the highest deviation in the direction $(1, 1)$. This eigenvector indicates that $x\!+\!x'$ has higher variance, while the variance of $x\!-\!x'$, with weights $(1, -1)$ in the orthogonal direction, is much lower. Thus, we know more about an image by knowing ``$x\!+\!x'\!=\!d_1$" than knowing ``$x\!-\!x'\!=\!d_2$", since the latter is already known to some extent by the smoothness assumption. We consider the fact ``neighboring pixels tend to be alike" 
as our prior-knowledge, and the constant value $d_1 \!=\! a x\! +\! a' x'$ as our post-knowledge, which is obtained from our reconstructed $\hat{\textbf{z}}_n^1$. When the vector $(a, a')$ is close to the first eigenvector $(1, 1)$ and orthogonal to $(1, -1)$ that corresponds to the prior-knowledge, we obtain the most information. 

In practice, for most existing CNNs, given the fact that $|\textbf{z}^1|<|\textbf{x}|$, where $|\cdot|$ denotes the length of vector, we will not be able to get a unique solution for all elements of input $\textbf{x}$. Instead, we want to set malicious parameters such that each element in $\textbf{z}^1$ reveals the sum of a group of neighboring pixels, and thus, we achieve the knowledge orthogonality and reconstruct a blurred image.




%
%
\vspace{-0.1cm}
\subsection{Convolutional Blocks Input Reconstruction} \label{subsec:conv}
\vspace{-0.1cm}
While there are many constraints to get the sum of neighboring pixels in the popular models, we first explain how we modify parameters to approximate the local sum. Specifically, we show our design on the VGG16~\cite{simonyan2014very} structure. \vspace{-0.3cm}
\begin{figure*}[ht]
	\centering
	\includegraphics[width=0.8\linewidth]{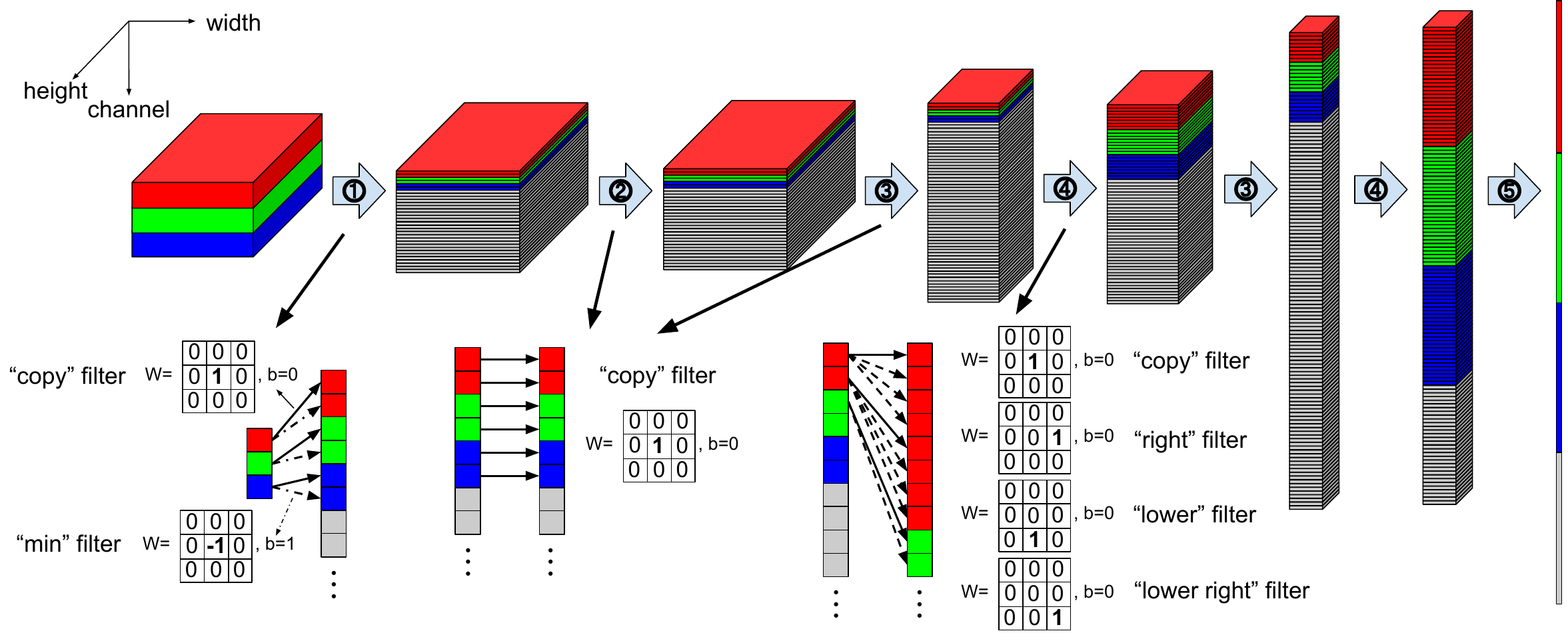}
 \vspace{-0.3cm}
	\caption{\small{Diagram of the naive approach of setting convolutional layer parameters to maximize knowledge orthogonality and obtain local sum of input $\textbf{x}$. The colored cuboids denote the input and output channels of each convolutional layer, and only the red, green and blue ones are considered. Layer \textcircled{\raisebox{-0.9pt}{1}} has both ``copy" filter and ``min" filter, and we only consider the first 6 channels in its output. Layer \textcircled{\raisebox{-0.9pt}{2}} copies the inputs, while layer \textcircled{\raisebox{-0.9pt}{3}} copies the input and also max-pools. Layer \textcircled{\raisebox{-0.9pt}{4}} divides each input channel into four output channels and shifts the features to corresponding directions. Layer \textcircled{\raisebox{-0.9pt}{5}} is a max-pooling layer with flattening. }
 }
	\label{fig:conv_orthogonal}
  \vspace{-0.5cm}
\end{figure*}
%
%
\vspace{-0.45cm}
\subsubsection{The Naive Approach} \label{subsubsec:conv_intuitive}
\vspace{-0.1cm}
In Fig.~\ref{fig:conv_orthogonal}, we sketch a naive approach to assign the malicious parameters on the VGG convolutional layers in a smaller scale for easier understanding. In this naive approach, the connected input and output channels have the convolutional filter parameters as assigned in the diagram, and all other parameters are set to zero. For the 3 RGB input channels, layer \textcircled{\raisebox{-0.9pt}{1}} divides the information of each channel into two different channels. The ``copy" filter copies the information, and the ``min" filter generates the reversed value (we assume each input element is a floating number with maximum value 1). Later on, in the input reconstruction part, we will use the former to get the local maximum, and the latter to get the local minimum. Layer \textcircled{\raisebox{-0.9pt}{2}} and layer \textcircled{\raisebox{-0.9pt}{3}} copy all the considered input channels, while layer \textcircled{\raisebox{-0.9pt}{3}} also has max-pooling with stride 2. Layer \textcircled{\raisebox{-0.9pt}{4}} divides each considered input channel into four channels. Each of the four filters copies the same information, or shifts the input features by one pixel to the right, down, and lower right. If there has been $i$ max-pooling layers before this layer, such filter will shift the considered region in the input by $2^i$ pixels. Finally, layer \textcircled{\raisebox{-0.9pt}{5}} is the max-pooling layer and flattens the output to a one-dimensional vector $\textbf{z}^1$ as the input of the FC classifier. We denote the tensor before flatten as $\textbf{z}^0$, which has the same set of elements but different shape than $\textbf{z}^1$.
\vspace{-0.3cm}
\subsubsection{Input Reconstruction} \label{subsubsec:conv_reconstruction}
\vspace{-0.15cm}
We now explain how to reconstruct input $\hat{\textbf{x}}_{n}$ from output $\hat{\textbf{z}}_n^1$ that is reconstructed from the FC classifier. For each sample, each element in ${\textbf{z}}^0$ denotes a local maximum or a local minimum within a certain region in input $\textbf{x}$. Assuming there are $I$-many 4-direction layers and $J$-many max-pooling layers with only ``copy" layer, the region size is $2^{I+J} \times 2^{I+J}$, and is much smaller than the receptive field. For VGG, the output ${\textbf{z}}^0$ has shape $7\!\times \!7\times\!512$, so each element ${\textbf{z}}^0[h, w, c]$ in the 6 channels, corresponding to only ``copy" filters, indicates the local maximum or minimum of a color in the considered region of the input $\textbf{x}$:
\vspace{-0.25cm}
\begin{multline} \label{eq:conv_recon_region}
\small{
\begin{aligned}
\textbf{R}[h, w, c] = \textbf{x}[&(h-1) 2^{I+J}+1 : h 2^{I+J}, \\
&(w-1) 2^{I+J}+1 : w 2^{I+J}, \lceil \frac{c}{2 \times 4^I} \rceil] ,
\end{aligned}
}
\vspace{-0.15cm}
\end{multline}
where the index $d_1:d_2$ denotes ``from $d_1$ to $d_2$", and $\lceil \cdot \rceil$ denotes rounding up. For each of the other elements ${\textbf{z}}^0[h, w, c]$ in the other channels, and for each $i^{th}$ 4-directional layer, the region is shifted to the right by $2^{i+J}$ pixels if $c$ corresponds to a ``right" filter or a ``lower right" filter, and we denote as indicator $\mathbf{1}_\text{r}(c, i) = 1$, otherwise 0. Also, this region is shifted down by $2^{i+J}$ pixels if $c$ corresponds to a ``lower" filter or a ``lower right" filter, and we denote as indicator $\mathbf{1}_\text{l}(c, i) = 1$, otherwise 0. Thus, the total shifted distance in input pixels is \vspace{-0.3cm}
\begin{equation} \label{eq:conv_recon_shift}
\small{
s_{\text{r}}(c) = \sum_i^I \mathbf{1}_\text{r}(c, i) 2^{i+J} , \quad
s_{\text{l}}(c) = \sum_i^I \mathbf{1}_\text{l}(c, i) 2^{i+J} ,
}
\vspace{-0.25cm}
\end{equation}
and the shifted considered region can be expressed as \vspace{-0.15cm}
\vspace{-0.2cm}
\begin{multline} \label{eq:conv_recon_region_shifted}
\small{
\begin{aligned}
\textbf{R}[h,& w, c] = \textbf{x}\Bigr[(h\!-\!1) 2^{I\!+\!J}\!+\!1\!+\!s_{\text{l}}(c) : h 2^{I\!+\!J}\!+\!s_{\text{l}}(c), \\
&(w\!-\!1) 2^{I\!+\!J}\!+\!1\!+\!s_{\text{r}}(c) : w 2^{I\!+\!J}\!+\!s_{\text{r}}(c), \lceil \frac{c}{2 \times 4^I} \rceil \Bigl] .
\vspace{-0.3cm}
\end{aligned}
}
\vspace{-0.3cm}
\end{multline}
Thus, each input element $\textbf{x}_n[h', w', c']$ of class $n$ is described by at most $2^{2I}$ maximum elements and $2^{2I}$ minimum elements in ${\textbf{z}}^0_n$. To get the tightest bound of the unknown element $\textbf{x}_n[h', w', c']$, we denote its estimated maximum and minimum as
\vspace{-0.15cm}
\begin{equation} \label{eq:conv_recon_max}\small
\operatorname*{max} \left( \textbf{x}_n[h', w', c'] \right) \stackrel{\text{def}}{=} \operatorname*{min}_{\substack{{\forall \{ c | \textbf{x}_n[h', w', c'] \in \textbf{R}[h, w, c], } \\ \lceil c / 4^I \rceil \text{ mod } 2 = 1 \}}} \left( {\textbf{z}}^0_n[h, w, c] \right) ,
\vspace{-0.7cm}
\end{equation}
\begin{equation} \label{eq:conv_recon_min}
\small
\operatorname*{min} \left( \textbf{x}_n[h', w', c'] \right) \stackrel{\text{def}}{=} \operatorname*{max}_{\substack{{\forall \{ c | \textbf{x}_n[h', w', c'] \in \textbf{R}[h, w, c], } \\ \lceil c/ 4^I \rceil \text{ mod } 2 = 0 \}}} \left( {\textbf{z}}^0_n[h, w, c] \right),
\vspace{-0.05cm}
\end{equation}
where $\text{mod}$ denotes the remainder. One way to estimate $\textbf{x}_n[h', w', c']$
is to take the average of (\ref{eq:conv_recon_max}) and (\ref{eq:conv_recon_min}) as our reconstructed input $\hat{\textbf{x}}_n$:
\vspace{-0.1cm}
\begin{equation} \label{eq:conv_recon_est}
\small{
\hat{\textbf{x}}_n[h'\!, w'\!, c']  \stackrel{\text{def}}{=} \frac{\operatorname*{max} \!\left( \textbf{x}_n[h'\!, w'\!, c'] \right) + \operatorname*{min} \!\left( \textbf{x}_n[h'\!, w'\!, c'] \right)}{2}.
}
\vspace{-0.15cm}
\end{equation}
Any reconstruction method using this malicious parameter setting should have at least $7\! \times\!7$ and at most $7\cdot 2^I \times7\cdot 2^I$ resolution. 
When the output $\hat{\textbf{z}}_n^1$ denotes a combination of samples from one class $n$, the reconstructed input $\hat{\textbf{x}}_n$ is also a combination of samples $\textbf{x}_k$. To improve the efficiency of utilizing 512 output channels of ${\textbf{z}}^0$, we pick $I=3$ so that $75\%$ of the channels are considered.
%
%
\vspace{-0.4cm}
\subsubsection{Inconspicuous Approach} \label{subsubsec:conv_inconspicuous}
\vspace{-0.2cm}
We now propose our inconspicuous approach to modify the network parameters so that it achieves the same reconstruction result as above while being more difficult to detect. 
For each layer, we denote the \emph{set of} considered input and output channels as $\mathcal{C}_{\text{in}}$ and 
$\mathcal{C}_{\text{out}}$, respectively. The indices of all channels are randomly shuffled, different from Fig. \ref{fig:conv_orthogonal}. 
If any of the filters in Fig.~\ref{fig:conv_orthogonal} is assigned to the network weight $\textbf{W}[c_{\text{out}}, c_{\text{in}}, :, :]$ and bias $\textbf{b}[c_{\text{out}}]$ where $c_{\text{in}}\in \mathcal{C}_{\text{in}}$ and $c_{\text{out}}\in \mathcal{C}_{\text{out}}$,
we denote this by the indicator function $\mathbf{1}(c_{\text{in}}, c_{\text{out}}) = 1$, and denote the corresponding parameters by $A_{\text{filter}}=\{W, b\}$. Otherwise, $\mathbf{1}(c_{\text{in}}, c_{\text{out}}) = 0$. We use $A_{\text{noise}}$ to denote i.i.d. noise parameters with Gaussian distribution
with zero mean and variance $\sigma^{2} \ll 1$. Therefore, we may assign each set of parameters in the network as
\vspace{-0.25cm}
\begin{align}\label{eq:conv_layer_param}
\small
A[c_{\text{in}}, c_{\text{out}}] = 
\begin{cases}
\beta A_{\text{filter}},\, \text{if } \mathbf{1}(c_{\text{in}}, c_{\text{out}}) = 1 \\
A_{\text{noise}},\, \text{else if } c_{\text{in}} \in \mathcal{C}_{\text{in}}, c_{\text{out}} \in \mathcal{C}_{\text{out}}  \\
A[c_{\text{in}}, c_{\text{out}}],\, \text{otherwise}
\end{cases} 
\vspace{-0.25cm}
\end{align}
where $\beta>0$ is a constant, which can be cancelled out during reconstruction. Since both $\mathcal{C}_{\text{in}}$ and $\mathcal{C}_{\text{out}}$ are only a small subset of all channels in most of the layers, the majority of the parameters are not modified, making it hard to detect. 
To better avoid being detected by any potential pattern-based detection, we can add i.i.d. noise to all modified parameters, or set smaller $I$ (i.e., set less 4-direction layers, and reconstruct with lower resolution). The overall process of MKOR on a VGG network is presented in 
Algorithm \ref{alg:malirecon}. \vspace{-0.15cm}
\begin{algorithm}
\caption{\footnotesize{Maximum Knowledge Orthogonality Reconstruction}}
\vspace{-0.05cm}
\label{alg:malirecon}
	\begin{algorithmic}
 \footnotesize
	    \State{Server sets the malicious parameters of FC layers according to (\ref{eq:1st_layer_weight})(\ref{eq:1st_layer_bias})(\ref{eq:latter_layer_weight})(\ref{eq:latter_layer_bias}), and sets those of conv layers according to (\ref{eq:conv_layer_param})}
	    \State{Server broadcasts modified neural network to each client $u$}
	    \State{\textbf{Clients execute:}}
	    \State{\textbf{for} client $u$ \textbf{do}}
            \State {$\quad\,$ Generate and upload the sum gradient $\sum_{k}^K \textbf{G}_k$ to server}
	    \State{\textbf{Server executes:}}
        \State{\textbf{for} each client $u$ \textbf{do}}
            \State{$\quad\,$ \textbf{for} each label $n$ \textbf{do}}
                \State{$\qquad\;\;$Solve $\hat{\textbf{z}}_n^1$ according to (\ref{eq:multi_layer_sol_intuitive})}
                \State{$\qquad\;\;$ $\hat{\textbf{z}}_n^0 = \text{reshape}(\hat{\textbf{z}}_n^1)$ }
                \State{$\qquad\;\;$ Estimate the input $\hat{\textbf{x}}_{n}$ according to (\ref{eq:conv_recon_est})}
         \vspace{-0.05cm}
	\end{algorithmic}
\end{algorithm}
\vspace{-0.1cm}
\subsection{Other Layers} \label{subsec:others}
\vspace{-0.1cm}
We now discuss the other types of layers that do not affect the overall function. For the dropout layer in FC classifier, we can set multiple parallel paths to each output label with the same functionality to increase the probability of one of the paths not being dropped.~As for the initial batch normalization and the batch-norm layers in convolutional layers, they only affect the magnitude of reconstruction, i.e., we will get a weighted input estimation $\gamma \hat{\textbf{x}}_{n}$, where $\gamma > 0$ depends on the batch-norm statistics.~We can calibrate the magnitude by modifying the histogram~\cite{gonzalez2009digital}, which may incur mean square error but does not affect human perception.

%
%
%
%
\vspace{-0.1cm}
\subsection{Applicability to Other Networks} \label{subsec:othernet}
\vspace{-0.1cm}
We can easily apply MKOR to other CNNs with similar architecture, such as LeNet~\cite{lecun1998gradient}. Two major differences of LeNet from VGG are the Sigmoid activation function instead of ReLU, and average pooling layers instead of max-pooling. Therefore, the ``min" filter is no longer necessary, and we can directly reconstruct the input $\hat{\textbf{x}}_{n}$ by inverting the Sigmoid activations in each convolutional layer from the features $\hat{\textbf{z}}_n^1$ of the first FC layer. 
\vspace{-0.2cm}
\section{Experiments}\label{sec:exp}
\vspace{-0.15cm}
We evaluate our MKOR on the MNIST dataset~\cite{deng2012mnist} with the LeNet5 network~\cite{lecun1998gradient}, and on higher resolution images, from CIFAR-100~\cite{krizhevsky2009learning} and ImageNet~\cite{deng2009imagenet} datasets, with the VGG16 network~\cite{simonyan2014very}. We perform comparison with baselines to illustrate the superior performance of MKOR. We consider the challenging cases of large batches, where the batch size is equal to or larger than the class number, 
without modifying the original network architecture. We consider the case of a single local update, where MKOR performs best. As stated in \cite{wang2021cooperative}, while more local updates allow higher system throughput, this incurs slightly higher error at convergence. 
Since no optimization or iterations are needed in MKOR, there are no additional hyper-parameters. MKOR can decode one batch with 100 CIFAR-100 images on VGG16 by a CPU within 2 min. In comparison, optimization-based methods require GPUs and take 8 hours to recover the same batch. 

\begin{figure}[b!]
 \vspace{-0.7cm}
	\centering
	\includegraphics[width=1\linewidth]{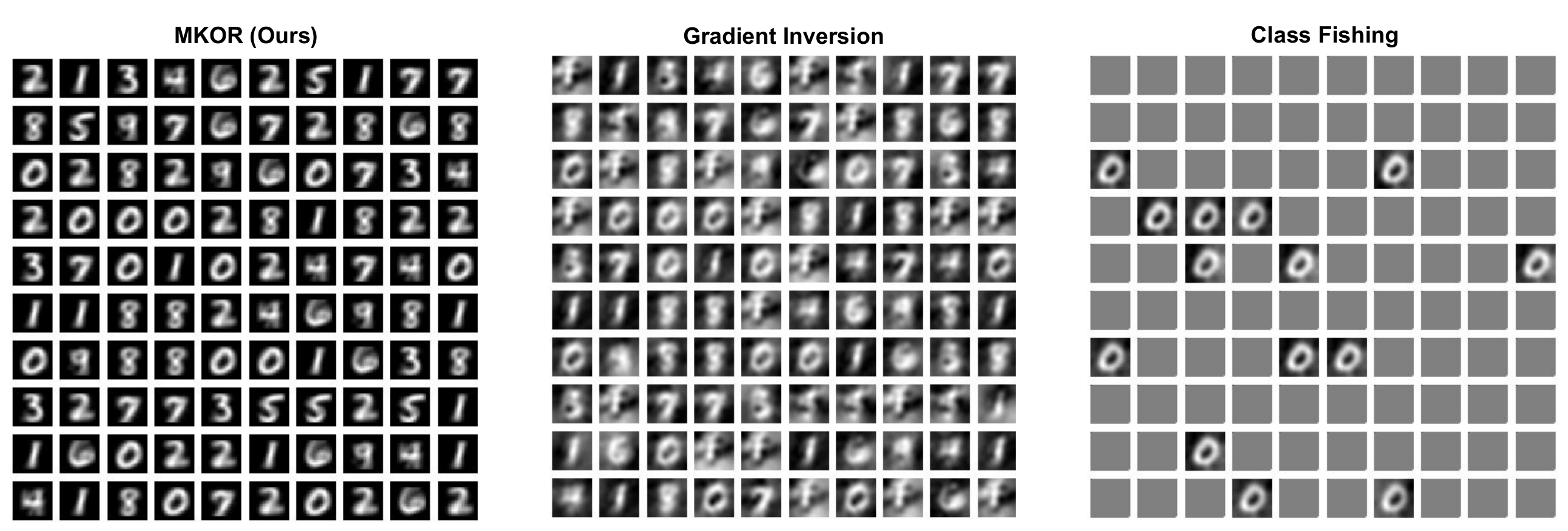}
 \vspace{-0.7cm}
	\caption{\small{Qualitative comparison of different input reconstruction methods on MNIST with original LeNet5 for a batch size of 100.}
 }
 \label{fig:MNIST_random100_originalLenet_only4}
  \vspace{-0.4cm}
\end{figure}

First, we reconstruct batches of 100 MNIST images with LeNet5, which is frequently used in previous work, where the batch size of 100 is larger than the number of classes (10). We compare MKOR with deep leakage from gradient (DLG)~\cite{zhu2019deep}, improved DLG (iDLG)~\cite{zhao2020idlg}, gradient inversion (GI)~\cite{geiping2020inverting}, and class fishing (CF)~\cite{wen2022fishing}, since these methods are more suitable for large batches than other existing methods. 
GI assumes an ``honest-but-curious" server, and CF uses malicious parameters without modifying the network architecture.~Yet, CF~\cite{wen2022fishing} can still be detected by checking for large number of zeros in the last layer.~In comparison, MKOR is not only suitable for large batches but also `inconspicuous' and hard to detect by clients as detailed in Sec.~\ref{subsubsec:dense_inconspicuous} and \ref{subsubsec:conv_inconspicuous}. Moreover, we notice that the majority of optimization-based attack methods, including DLG, iDLG, GI and CF, delete some or all of the pooling layers for better attack performance. 
We also use LeNet as was done in most of the baselines. We perform comparison with both original LeNet and modified LeNet (i.e., use stride=2 in conv layers instead of avg pooling layers). In Tab.~\ref{tab:performance_random_bs100}, a quantitative comparison is presented in terms of the structural similarity index measure (SSIM) and peak signal-to-noise ratio (PSNR) between the original and the reconstructed image. We note that since CF can reconstruct the image of one class only, the average SSIM and average PSNR are not available.~Tab.~\ref{tab:performance_random_bs100} shows the result of 10 random batches. Since the batch size of 100 is greater than the number of classes, images are randomly selected and there are repeated classes.~Thus, each cell in Tab.~\ref{tab:performance_random_bs100} is the average of 1000 images. 
For a qualitative comparison, we present a reconstructed batch of images by MKOR, GI and CF using original LeNet in Fig.~\ref{fig:MNIST_random100_originalLenet_only4}.~The results of DLG and iDLG, and the images reconstructed with the modified LeNet can be found in the Suppl. material with larger figures. From Tab.~\ref{tab:performance_random_bs100} and Fig.~\ref{fig:MNIST_random100_originalLenet_only4}, MKOR outperforms all the baselines, and from Tab.~\ref{tab:performance_random_bs100} and Fig.~2 in the Suppl. file, MKOR has even better performance when modified LeNet is used, where modification does not involve using a very large FC layer. 

Since optimization-based methods rely heavily on the gradient of convolutional layers, they do not perform well on large batches.~As explained in Sec.~\ref{subsec:necessary}, reconstruction from large batches should focus more on extracted features.
\begin{table}[tb!]
\vspace{-0.25cm}
    \small
	\begin{center}
	    \caption{\small{Comparison of different reconstruction methods on MNIST with LeNet5 for 10 random batches with batch size 100.}}
      \vspace{-0.25cm}
     \resizebox{0.82\linewidth}{!}{%
		\begin{tabular}{ | l |  l | l | } \hline
			\vtop{\hbox{\strut \emph{max}, avg SSIM}\hbox{\strut \emph{max}, avg PSNR}} & Original LeNet & Modified LeNet \\   \hline
            DLG & \emph{0.13}, 0.02, \emph{2.00}, 0.52 & \emph{0.08}, 0.01, \emph{1.21}, 0.20 \\   \hline
            iDLG & \emph{0.11}, 0.03, \emph{1.28}, 0.18 & \emph{0.14}, 0.04, \emph{1.29}, 0.26 \\   \hline
            GI & \emph{0.30}, 0.13, \emph{15.64}, 11.44 & \emph{0.41}, 0.18, \emph{15.96}, 11.48 \\   \hline
            CF & \emph{0.21}, —, \emph{14.63}, — & \emph{0.42}, —, \emph{15.95}, — \\   \hline
            MKOR (ours) & \textbf{0.54}, \textbf{0.27}, \textbf{17.06}, \textbf{12.79} & \textbf{0.69}, \textbf{0.38}, \textbf{18.72}, \textbf{13.53} \\   \hline
		\end{tabular}
  }
		\label{tab:performance_random_bs100}
	\end{center}
 \vspace{-1cm}
\end{table}
\vspace{-0.4cm}
In Tab.~2 of the Suppl. file, we consider additive Gaussian noise on gradients as used in differential privacy \cite{abadi2016deep}. MKOR still outperforms all the baselines under high noise, which provides higher level of privacy protection. We note that differential privacy and local differential privacy \cite{bhowmick2018protection} fail under such attack, because these works do not take the neural network structure and its parameters into account. More discussion is provided in the Suppl. file. 
Since MKOR does not depend on image distribution, transformation-based defenses, such as \cite{gao2021privacy}, do not work either, and MKOR can still recover the transformed input. 
In Tab.~3 of the Suppl. file, we consider Soteria \cite{sun2021soteria}, which  optimizes the perturbation such that the reconstructed data is severely degraded. As seen, while Soteria is a general defense method not focusing on a specific attack, it is not as effective on MKOR as the other attacks.
Please see the Suppl. material for additional details, the images showing the reconstruction under Gaussian noise and Soteria defense, and single large samples for reconstruction detail. 

\begin{figure}
 \vspace{-0.3cm}
	\centering
	\includegraphics[width=0.8\linewidth]{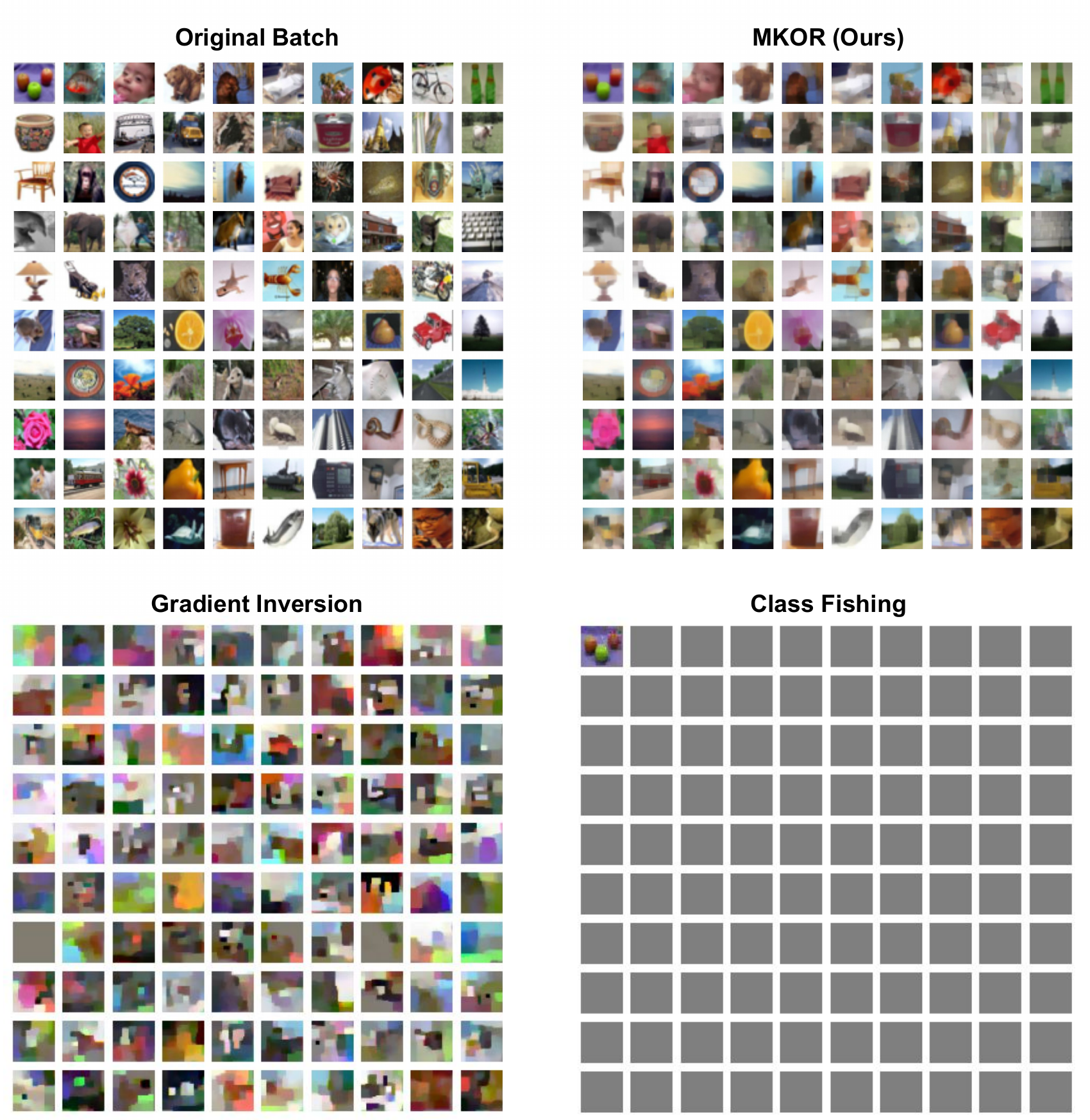}
 \vspace{-0.3cm}
	\caption{\small{Qualitative comparison between different input reconstruction methods on CIFAR-100 for a batch size of 100.}
 }
	\label{fig:outputs}
  \vspace{-0.2cm}
\end{figure}

\begin{figure}[h!]
 \vspace{-0.cm}
	\centering
	\includegraphics[width=0.8\linewidth]{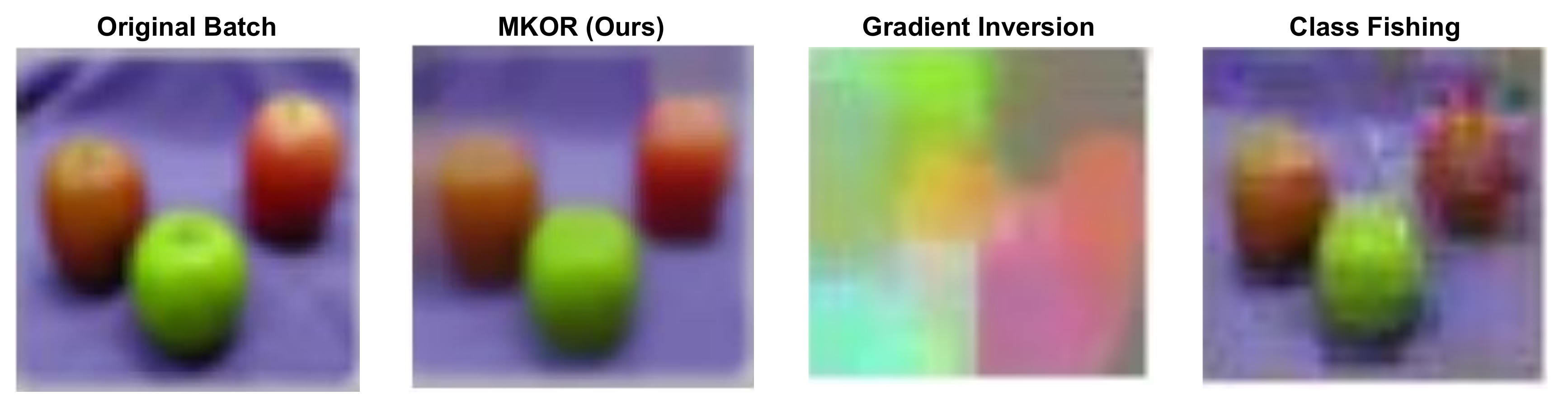}
 \vspace{-0.4cm}
	\caption{\footnotesize{Larger plots of samples from Fig.~\ref{fig:outputs} for reconstruction details.}
 }
\label{fig:OneLine_SingleSample_outputs_copy_gradinver_input}
  \vspace{-0.5cm}
\end{figure}

Since GI and CF (although for a single class) performed better in the MNIST experiment with large batch size, we compare MKOR with GI and CF
on CIFAR-100 dataset~\cite{krizhevsky2009learning} with VGG16 network~\cite{simonyan2014very}. We use 
their experiment setting of unique batches, where each image is a unique sample, being the only one representing its class.~The batch size $K\!\!\!=\!\!\!100$, which is the same as the number of classes. The extracted features $\textbf{z}^0$ from VGG conv. layers are of shape 7$\times$7$\times$512.~If an arbitrary function is used (instead of a CNN), we can reconstruct images with 91$\times$91 resolution.~Yet, due to the practical limit of the CNN structure, the function is not arbitrary, so our post-knowledge is limited with more information loss. Here, we show the results obtained by Alg.~ \ref{alg:malirecon}. Since we pick the number of 4-direction layers as $I\!\!=\!\!3$, the expected resolution of MKOR is between $7 \times 7$ and $7 \cdot 2^I \times 7 \cdot 2^I = 56 \times 56$. In Tab.~\ref{tab:performance}, we show the quantitative comparison. With a batch size of 100, MKOR outperforms both GI and CF.~We present a reconstructed batch of images in Fig.~\ref{fig:outputs} 
and single reconstructed samples for a closer look in Fig.~\ref{fig:OneLine_SingleSample_outputs_copy_gradinver_input} (please refer to Suppl.~material for additional and larger images).
As can be seen, CF~\cite{wen2022fishing} can only reconstruct one image per batch, and the images reconstructed by GI~\cite{geiping2020inverting} are distorted. In contrast, with MKOR, most images in the batch are better recognizable. Moreover, while modifying the network architecture does not make a difference on GI and CF performance, MKOR with a modified network can accurately reconstruct an arbitrary dataset when there is no image with repeating class. For such unique batch from CIFAR-100, we have [max SSIM, avg SSIM, max PSNR, avg PSNR] = [1.000, 1.000, 163.557, 156.650]. The plots showing the reconstruction of the same batch size $K\!\!\!=\!\!\!100$, but with less number of classes $N$ can be found in the Suppl. material. 

\begin{table}[t!]
    \small
	\begin{center}
	    \caption{\small{Comparison of different input reconstruction methods on CIFAR-100.~Each data point shows the mean and standard deviation over the same $U=10$ set of batches.}}
      \vspace{-0.32cm}
     \resizebox{0.95\linewidth}{!}{%
		\begin{tabular}{ | l |  l | l | l | l | } \hline
			Methods & \vtop{\hbox{\strut Maximum}\hbox{\strut SSIM ($\uparrow$)}}  & \vtop{\hbox{\strut Average}\hbox{\strut SSIM ($\uparrow$)}} & \vtop{\hbox{\strut Maximum}\hbox{\strut PSNR ($\uparrow$)}} & \vtop{\hbox{\strut Average}\hbox{\strut PSNR ($\uparrow$)}}  \\   \hline
            GI  & 0.52 $\pm$ 0.04 & 0.37 $\pm$ 0.01 & 17.51 $\pm$ 1.61 & 11.2 $\pm$ 0.36 \\  \hline
            CF & 0.81 $\pm$ 0.10 & - & 21.56 $\pm$ 4.65 & - \\ \hline
            MKOR (Ours) & \textbf{0.98 $\pm$ 0.01} & \textbf{0.87 $\pm$ 0.00} & \textbf{33.85 $\pm$ 2.07} & \textbf{24.14 $\pm$ 0.28} \\  \hline
		\end{tabular}
  }
		\label{tab:performance}
	\end{center}
 \vspace{-0.8cm}
\end{table}
\begin{table}[htb!]
\vspace{-0.4cm}
    \small
	\begin{center}
	    \caption{\small{Performance of MKOR with VGG16 on ImageNet of batch size 1000}}
      \vspace{-0.32cm}
     \resizebox{1\linewidth}{!}{%
		\begin{tabular}{ | l |  l | l | } \hline
			\vtop{\hbox{\strut \emph{max}, avg SSIM}\hbox{\strut \emph{max}, avg PSNR}} & \vtop{\hbox{\strut Original VGG16}\hbox{\strut BS=1000}} & \vtop{\hbox{\strut Modified network}\hbox{\strut BS=1000}} \\   \hline
            unique batch, no noise & \emph{0.93}, 0.48, \emph{30.12}, 17.88 & \emph{1.00}, 1.00, \emph{inf}, inf \\   \hline
            unique batch, Gaussian $(10^{-1})$ & \emph{0.92}, 0.45, \emph{29.99}, 17.50 & \emph{0.97}, 0.80, \emph{31.24}, 31.00 \\   \hline
            random batch, no noise & \emph{0.90}, 0.42, \emph{30.05}, 15.05 & \emph{1.00}, 0.72, \emph{inf}, inf \\   \hline
            random batch, Gaussian $(10^{-1})$ & \emph{0.88}, 0.41, \emph{26.44}, 14.80 & \emph{0.88}, 0.52, \emph{28.37}, 17.63 \\   \hline
		\end{tabular}
  }
		\label{tab:ImageNet_bs1000}
	\end{center}
 \vspace{-0.5cm}
\end{table}

In Tab.~\ref{tab:ImageNet_bs1000}, we evaluate MKOR with larger batch (1000 in this case) and larger image resolution (224$\times$224 on ImageNet vs. 32$\times$32 with CIFAR100) on VGG16 and modified VGG16 (conv with stride=2 instead of pooling layers) with and without Gaussian noise with normalized standard deviation $(10^{-1})$.~We test MKOR on 10 unique batches and 10 random batches of batch size 1000. While most works do not attempt to use a batch size of 1000 on ImageNet, MKOR is able to maintain excellent performance, and we have not seen similar works on such large batches with unmodified benchmark models. It should be noted that optimization-based methods on this batch size of high-resolution images require prohibitively large memory and long runtime. 
The images showing the reconstruction with this setting can be found in the Suppl.~material.

\vspace{-0.2cm}
\section{Conclusion}\label{sec:con}
\vspace{-0.2cm}
We have introduced the maximum knowledge orthogonality reconstruction (MKOR) as a novel federated learning (FL) input reconstruction method, 
which can be used with large batches.~Our approach maliciously sets the parameters of fully-connected layers to accurately reconstruct the input features, and of convolutional layers to maximize the orthogonality between prior and post knowledge.~For both cases, we have proposed an inconspicuous approach 
to avoid being detected by skeptical clients.~Experiments have demonstrated that MKOR outperforms other baselines on large batches of the MNIST dataset with LeNet model, and on CIFAR-100 and ImageNet datasets with a VGG16 model by providing better reconstructed images both qualitatively and quantitatively. Our results encourage further research on the protection of data privacy in FL.

{\small
\bibliographystyle{ieee_fullname}
\bibliography{ref.bib}
}

\clearpage
\section{Additional Proof}
In this section, we provide detailed explanations, interpretations, and proof of the mathematical content for comprehensive insight. We list all the notations used in the paper in Table~\ref{tab:notation}.

We derive \textbf{Eqn. (2)} included in our main paper as shown below, where $O(\frac{1}{K})$ denotes order $K^{-1}$ complexity.  
\begin{multline*} \label{eq:finite_leakage_converge}
\small{
\begin{aligned}
& \quad\  H(\textbf{g}_1) - H\left(\textbf{g}_1 | \sum_{k=1}^{K} \textbf{g}_k\right) \\
& = H\left(\sum_{k=1}^{K} \textbf{g}_k\right) - H\left(\sum_{k=1}^{K} \textbf{g}_k | \textbf{g}_1\right)  \\
& = H\left(\sum_{k=1}^{K} \textbf{g}_k\right) -  H\left(\sum_{k=2}^{K} \textbf{g}_k\right) \\
& = \frac{1}{2}\log(2\pi K \sigma_g^2)+\frac{1}{2} - \frac{1}{2}\log(2\pi (K-1) \sigma_g^2)-\frac{1}{2} \\
& = \frac{1}{2}\log\left( 1+\frac{1}{K-1} \right) \\
& = \frac{1}{2(K-1)} + \frac{1}{2} \left( \frac{1}{2(K-1)} \right) ^2 + \ldots \\
&= O\left( \frac{1}{K} \right). \\
\end{aligned}
}
\end{multline*}

\begin{table}
    \small
	\begin{center}
	    \caption{Notation}
     \vspace{-0.4cm}
		\begin{tabular}{  p{24mm} p{57mm}  }\hline
            \underline{FL Parameters:} &  Section 2, 3.1 \\
			$U$, $u$ & Total number and index of clients \\ 
            $K$, $k$ & Batch size and local sample index \\
            $\textbf{x}_{k}$, $y_{k}$ & The $k^{th}$ input image and label \\
            $\textbf{A}$ & Network parameters (weights and biases) \\
            $\mathcal{L}(\textbf{x}, y \, | \, \textbf{A})$ & Loss function \\
			$\textbf{G}_k$ & Gradient of the $k^{th}$ image \\   
            $N$ & Total number of labels \\
            $n$ & Index of node in the corresponding layer \\
			\hline
            \underline{MKOR - FC:} & Section 3.2 \\
            $\textbf{p}=\{p_1, \ldots, p_{N}\}$ & Local sample distribution over classes \\
            $L$, $l$ & Total number of layers and layer index \\
            $N^l$ & Total number of input nodes \\
            $\textbf{W}^l$, $\textbf{b}^l$ & Weight and bias at FC layer $l$ \\
            $\textbf{z}_k^l$ & Input from sample $k$ at FC layer $l$ \\
            $\textbf{y}_k$ & Classifier output from sample $k$ \\
            $\hat{\textbf{z}}_n^1$ & Reconstructed input of label $n$ \\
			\hline
            \underline{MKOR - conv:} & Section 3.4 \\
            $\textbf{z}^0$ & Output of convolutional layers before flatten \\
            $I$, $i$ & Total number and index of 4-direction layers \\
            $J$, $j$ & Total number and index of max-pooling layer with only ``copy" layer \\
            $h$, $w$, $c$ & Height, width and channel index \\
            $\textbf{R}[h, w, c]$ & Considered region in $\textbf{x}$ for $\textbf{z}^0[h, w, c]$ \\
            $\hat{\textbf{x}}_n$ & Reconstructed input image of label $n$ \\
            \hline
		\end{tabular}
		\label{tab:notation}
	\end{center}
     \vspace{-0.9cm}
\end{table}

We first rewrite  \textbf{Eqn. (4)} from the main paper as below: 
\begin{equation*}
\footnotesize{
\frac{\partial \mathcal{L}(\textbf{x}_k, y_k \, | \, \textbf{A})}{\partial \textbf{W}^1} = \frac{\partial \mathcal{L}(\textbf{x}_k, y_k \, | \, \textbf{A})}{\partial \textbf{y}_k} \frac{\partial \textbf{y}_k}{\partial \textbf{W}^1} = \frac{\partial \mathcal{L}(\textbf{x}_k, y_k \, | \, \textbf{A})}{\partial \textbf{b}^1} {\textbf{z}_k^1}^\text{T}. 
}
\end{equation*}
We note that the first equality is derived from the chain rule. To prove the second equality, we first clarify that an FC classifier with only one layer has no activation function, therefore $\textbf{y}_k = \textbf{W}^1 \textbf{z}_k^1 + \textbf{b}^1$. Hence, $\frac{\partial \mathcal{L}(\textbf{x}_k, y_k \, | \, \textbf{A})}{\partial \textbf{y}_k} = \frac{\partial \mathcal{L}(\textbf{x}_k, y_k \, | \, \textbf{A})}{\partial \textbf{b}^1}$. Additionally, the partial derivative $\frac{\partial \textbf{y}_k}{\partial \textbf{W}^1}$ is simply the transpose of the multiplier ${\textbf{z}_k^1}^\text{T}$. 
For the multiple layer case as we claimed in \textbf{Eqn. (9)}, we may consider a dummy vector $\textbf{z}'_k$ before activation $f$, i.e. $\textbf{z}^2_k = f(\textbf{z}'_k) = f(\textbf{W}^1 \textbf{z}_k^1 + \textbf{b}^1)$. Thus, we can substitute $\textbf{z}'_k$ for $\textbf{y}_k$ in the analysis above, and get the same result as shown in Eqn. (9).

For the special case of an FC classifier without bias term, we have to set every connected weight in Fig. 1 of the main paper to the same positive value. Thus, for each input sample, all elements of the vector $\frac{\partial \mathcal{L}(\textbf{x}_k, y_k \, | \, \textbf{A})}{\partial \textbf{y}_k}$ will have the same value. Therefore, although we do not have this vector indicating the magnitude of the reconstructed image, we still have a rough estimation that is uniformly brighter or darker than the original input with the same ratio between pixels, and it is usually sufficient to identify all key features. Nevertheless, for the majority of benchmark models, we usually have the bias term. 

The intuition behind \textbf{Equations (10), (11), (12) and (13)} is to decouple one node for each of the $N$ labels in the first $2N$ nodes in the output of the first FC layer $\textbf{z}^2$. Such decoupling is similar to the naive case shown in Fig. 1, except that each output node is coupled to two different nodes in the output in the second layer. As shown in Eqn. (10), for each pair of neighboring nodes of $\textbf{z}^2$, the weight in the first layer to one node is the opposite of the weight to the other node. Therefore in the forward-propagation, one node with negative output will be zeroed out by ReLU activation while another with positive output will be kept. If we alternatively consider activation that does not zero out negative values such as Sigmoid, we can use only one line of nodes for each label. In the case where activation zeroes out negative values such as ReLU, we note in Eqn. (11) that the bias in the first layer to each decoupled node should be modified accordingly, to avoid zeroing out necessary forward-propagation paths. 

\textbf{Eqn. (12)} specifies the weights in the second FC layer and the following FC layers to ensure that the target $N$ nodes in the first $2N$ nodes of $\textbf{z}^2$ are decoupled. In the second FC layer, we merge every two neighboring nodes in the first $2N$ nodes of the input $\textbf{z}^2$ to a single node in the first $N$ nodes of the output $\textbf{z}^3$. According to the pair of opposite weights in the first layer, for each pair of nodes in the first $2N$ nodes of $\textbf{z}^2$, there will be one node with positive value and the other with zero value after ReLU activation. Therefore, we set the weight of the merging path to positive values and the rest to minor Gaussian noise, so that the back-propagation path will go back to one of the decoupled pair of nodes with positive output. For the majority of FC layers, $N \ll N^l$ and such modification can be hardly detected. For the following layers (i.e., 3rd, 4th, etc), we keep a single positive weight for each decoupled node until the output. \textbf{Eqn. (13)} specifies that the bias to the decoupled nodes is non-negative in order to prevent the output being zeroed out by ReLU activation.

To explain the convolutional block reconstruction, we first illustrate how the forward-propagation in \textbf{Fig. 2} works. We denote the input nodes as $\textbf{x}$, and the output nodes of each convolutional layers as $\textbf{x}^1$, $\textbf{x}^2$, etc. For a typical RGB image, we have three input channels in $\textbf{x}$. After the forward-propagation, the first layer of type \textcircled{\raisebox{-0.9pt}{1}} divides the information of each input channel to a positive copy and a negative copy. That is to say, channel 1 in $\textbf{x}$ is exactly copied to channel 1 in $\textbf{x}^1$ which is called a ``copy" filter, and a negative copy $1 - \textbf{x}$ is copied to channel 2 in $\textbf{x}^1$ which is called a ``min" filter. The benefit of such division is to keep both maximum values and minimum values after the max-pooling layer. Similarly, channel 2 in $\textbf{x}$ is divided into channels 3 and 4 in $\textbf{x}^1$, and channel 3 in $\textbf{x}$ is divided into channels 5 and 6 in $\textbf{x}^1$. Therefore, all the information in $\textbf{x}$ is kept in the first 6 channels in $\textbf{x}^1$, and we do not care about the rest of the channels in $\textbf{x}^1$. The second layer of type \textcircled{\raisebox{-0.9pt}{2}} copies the 6 channels to the first 6 channels in $\textbf{x}^2$. The third layer of type \textcircled{\raisebox{-0.9pt}{3}} also copies the 6 channels to $\textbf{x}^3$, but it also has a $2 \times 2$ max-pooling. As a result, each node in channel 1 of $\textbf{x}^3$ denotes the maximum value of a $2 \times 2$ area in channel 1 of $\textbf{x}$, and each node in channel 2 of $\textbf{x}^3$ denotes the minimum value of a $2 \times 2$ area in channel 1 of $\textbf{x}$. Similarly, channels 3 to 6 of $\textbf{x}^3$ represents the maximum and minimum values of channels 2 and 3 of $\textbf{x}$. While we lost a half of the width and a half of the height in $\textbf{x}^3$ and the amount of information is reduced to $\frac{1}{4}$ in each channel, we also have many more channels to utilize. To store the input information in more different channels, we in the fourth layer of type \textcircled{\raisebox{-0.9pt}{4}} divide each channel in $\textbf{x}^3$ into 4 different channels in $\textbf{x}^4$ in 4 different directions. Specifically, the output channel with ``copy" filter focuses on the same area of input $\textbf{x}^3$, and the output channel with ``right" filter, ``lower" filter or ``lower right" filter focuses on a switched area of the input $\textbf{x}^3$. Thus, these 4 channels represent 4 different areas in the input. These areas are similar, but slightly switched. The fifth layer and the sixth layer have the same function as the third layer and the fourth layer, and the considered region is further switched. Finally, the last layer max-pools and flattens the output. Given the two switches with layer type \textcircled{\raisebox{-0.9pt}{4}}, each output node that corresponds to the considered channels maps to a different considered region in the input $\textbf{x}$, and our input reconstruction relies on these switches.

With the aforementioned idea of Fig. 2 in mind, we then explain the rest of the equations. In \textbf{Eqn. (14)}, we consider the 6 output channels that are not shifted, i.e., all corresponding filters are ``copy" filters. Each element in these 6 output channels corresponds to a certain ``considered region" in the input, which is defined in Eqn. (14). For the rest of the output channels with at least one switch, i.e., corresponding to at least one of the ``right" filter, ``lower" filter or ``lower right" filter, we define the shifted distance at the input image $\textbf{x}$ in \textbf{Eqn. (15)}. Given the shifted distance, we express the shifted considered region for arbitrary output channel in \textbf{Eqn. (16)}, which is a generalized version of Eqn. (14). On the one hand, a half of the channels experienced a ``copy" filter at the first layer thus describing the maximum value of the considered region, while the others experienced a ``min" filter at the first layer thus describing the minimum value of the considered region. On the other hand, since each output node describes a considered region in the input, each node of the input is also described by multiple output nodes. Therefore, in \textbf{Eqn. (17)}, the upper bound of an input node is defined as the minimum of the maximum describers. In \textbf{Eqn. (18)}, the lower bound of an input node is defined as the maximum of the minimum describers. Without further information, we estimate the input node by the average between the upper bound and the lower bound as in \textbf{Eqn. (19)}. For the inconspicuous approach, the indices of the channels should be randomly shuffled. In \textbf{Eqn. (20)}, we can multiply the weight and bias of the filters in each layer by a different positive factor $\beta$, and simply divide by different $\beta$s during reconstruction.

\section{Additional Examples}
In this section, we provide comparison between experimental settings and additional examples with larger images. 

In our inconspicuous reconstruction approaches, we proposed multiple strategies such as reshuffling of indices and magnitude manipulation to reduce the detectability by the clients. Furthermore, even if we only focus on the ratio of the modified parameters, MKOR is still much more inconspicuous than the existing approaches. In MKOR for VGG16, the layer with highest ratio of modification is the last conv layer. It has 2359808 parameters, and 332160 parameters are modified, so the peak modification ratio is 14\%. Furthermore, the overall modification ratio for all layers in VGG16 is less than 1\%, and every image generates gradient of similar magnitude. In contrast, class fishing \cite{wen2022fishing} sets 99.9\% of parameters in the last layer to zero, and input images from 99.9\% of the labels do not generate any gradient at all (i.e., all of them have zero gradient). Other works assuming a malicious server either insert FC layer with tremendous number of input nodes (about $10^8$), or even modify the federated learning architecture. Both cases are much more easily detectable than MKOR, and please refer to Section 2 in the paper for details. Also, different from some malicious server attacks, the performance of MKOR does not depend on any unmodified parameter, so there is no dependency on the accuracy or convergence of the network.

\begin{figure*}[h]
	\centering
	\includegraphics[width=0.75\linewidth]{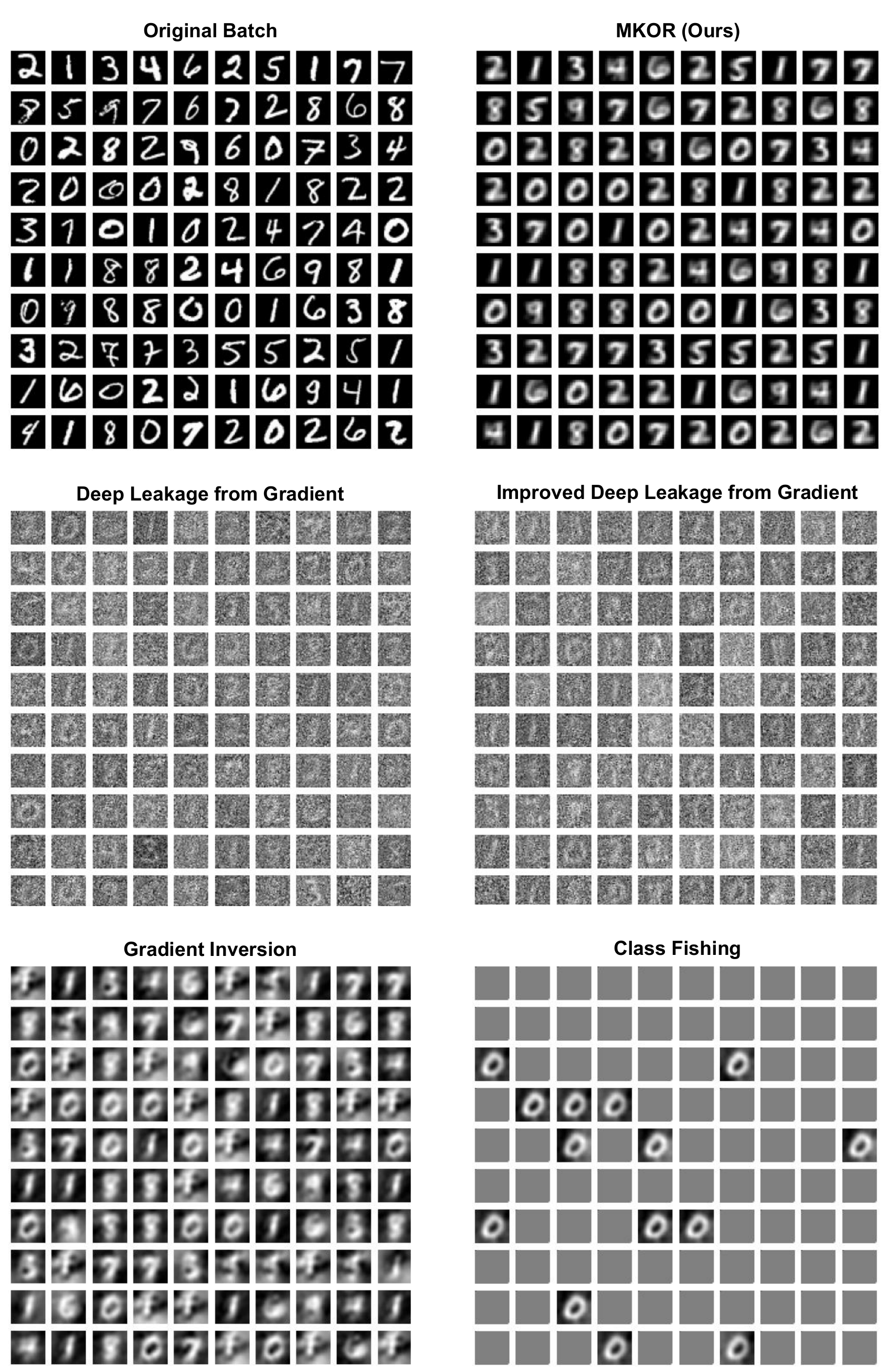}
 \vspace{-0.3cm}
	\caption{Qualitative comparison between different input reconstruction methods on MNIST with original LeNet5 for a batch size of 100.
 }
	\label{fig:MNIST_random100_originalLenet}
  \vspace{-0.4cm}
\end{figure*}

\begin{figure*}[h]
\centering
	\includegraphics[width=0.75\linewidth]{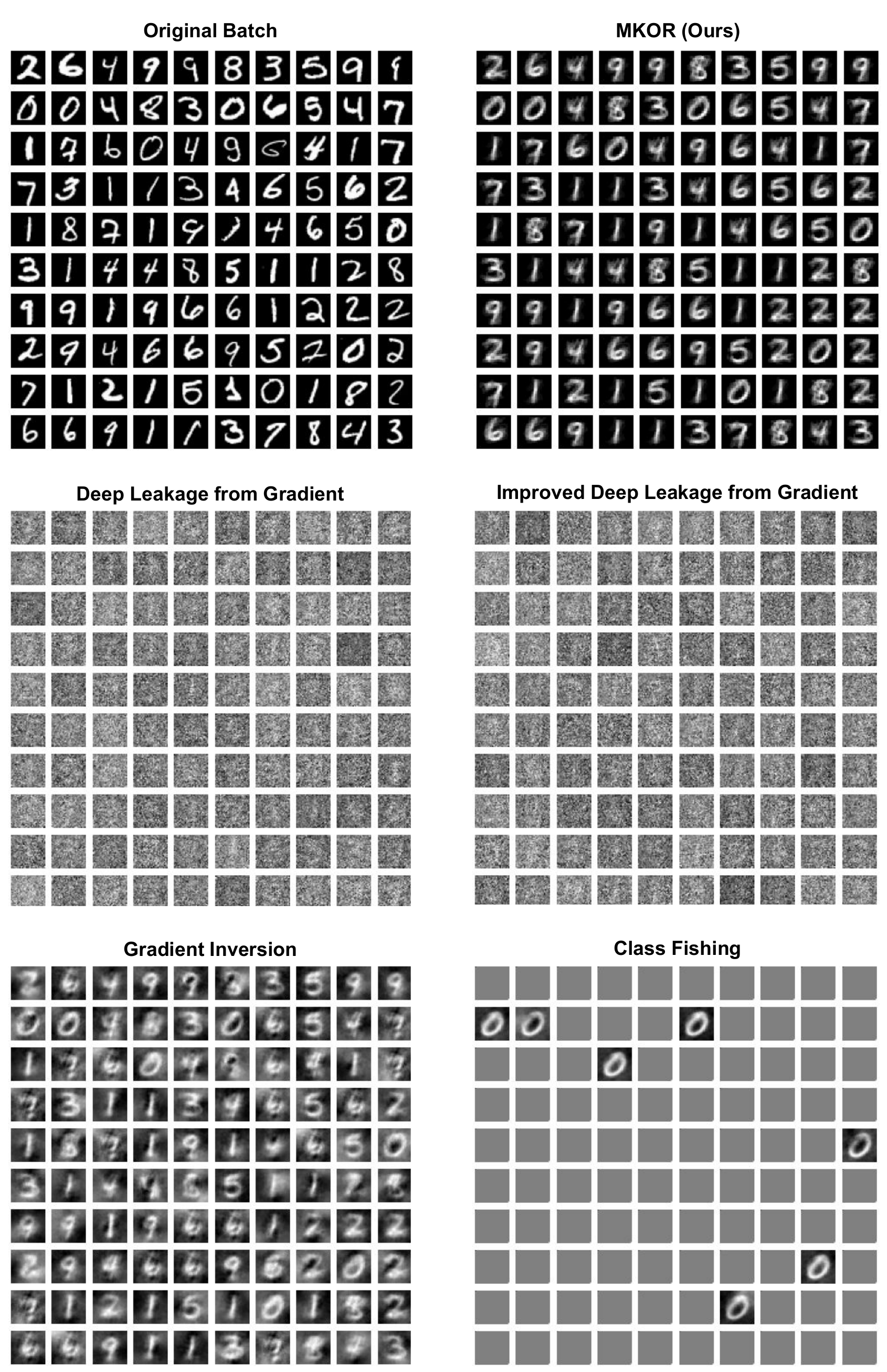}
 \vspace{-0.3cm}
	\caption{Qualitative comparison between different input reconstruction methods on MNIST with modified LeNet for a batch size of 100.
 }
	\label{fig:MNIST_random100_modifiedLenet}
  \vspace{-0.4cm}
\end{figure*}

In Fig.~\ref{fig:MNIST_random100_originalLenet} and Fig.~\ref{fig:MNIST_random100_modifiedLenet}, we consider reconstructing a batch with 100 MNIST images via original LeNet and modified LeNet, respectively, with MKOR, and compare with deep leakage from gradient (DLG)~\cite{zhu2019deep}, improved DLG (iDLG)~\cite{zhao2020idlg}, gradient inversion (GI), and class fishing~\cite{wen2022fishing}. 

\begin{figure*}[h]
\centering
	\includegraphics[width=0.75\linewidth]{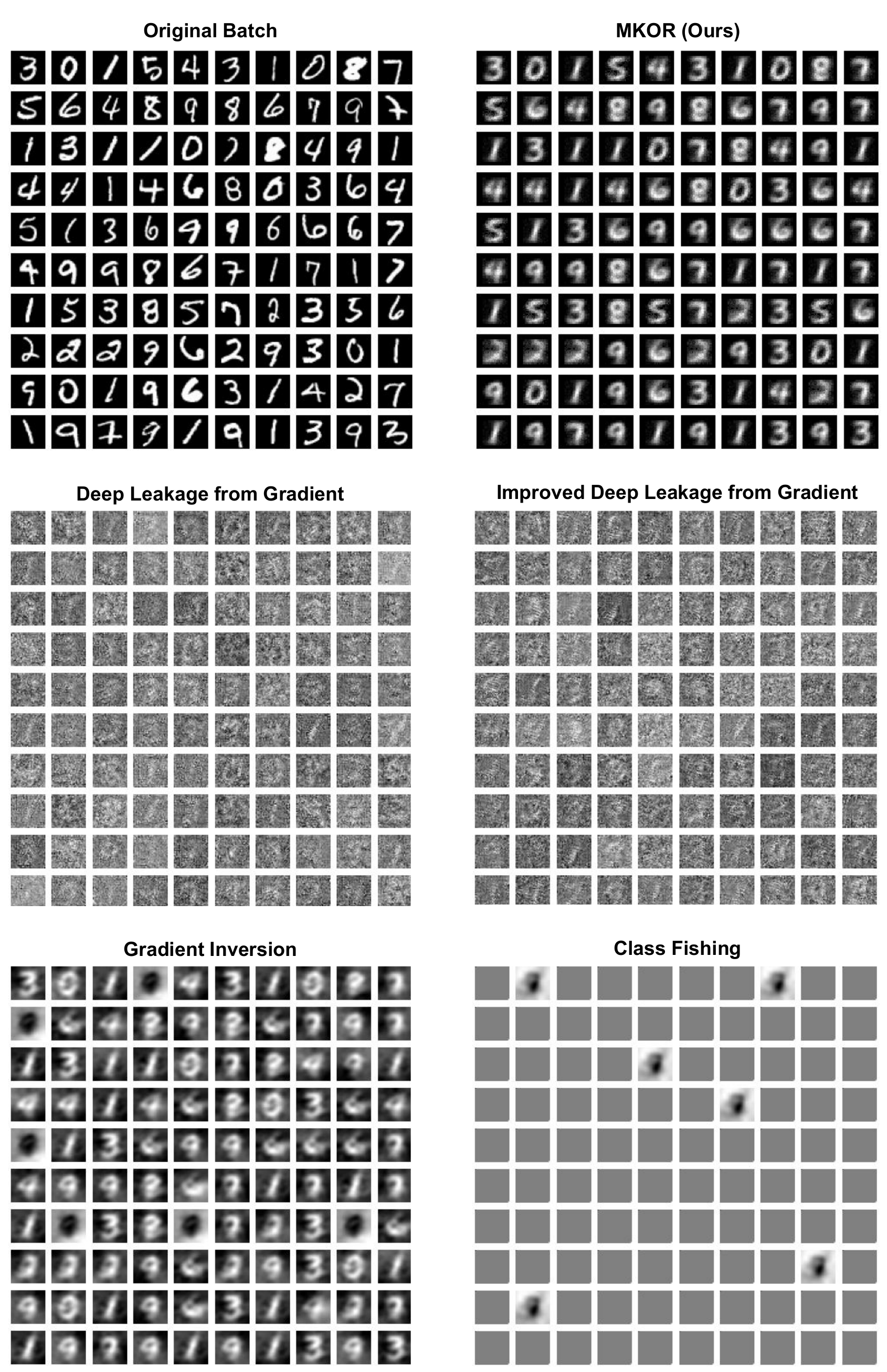}
 \vspace{-0.3cm}
	\caption{Qualitative comparison between different input reconstruction methods under Gaussian noise on MNIST with original LeNet5 for a batch size of 100.
 }
	\label{fig:MNIST_random100_originalLenet_Gaussian}
\end{figure*}

\begin{figure*}[h]
\centering
	\includegraphics[width=0.75\linewidth]{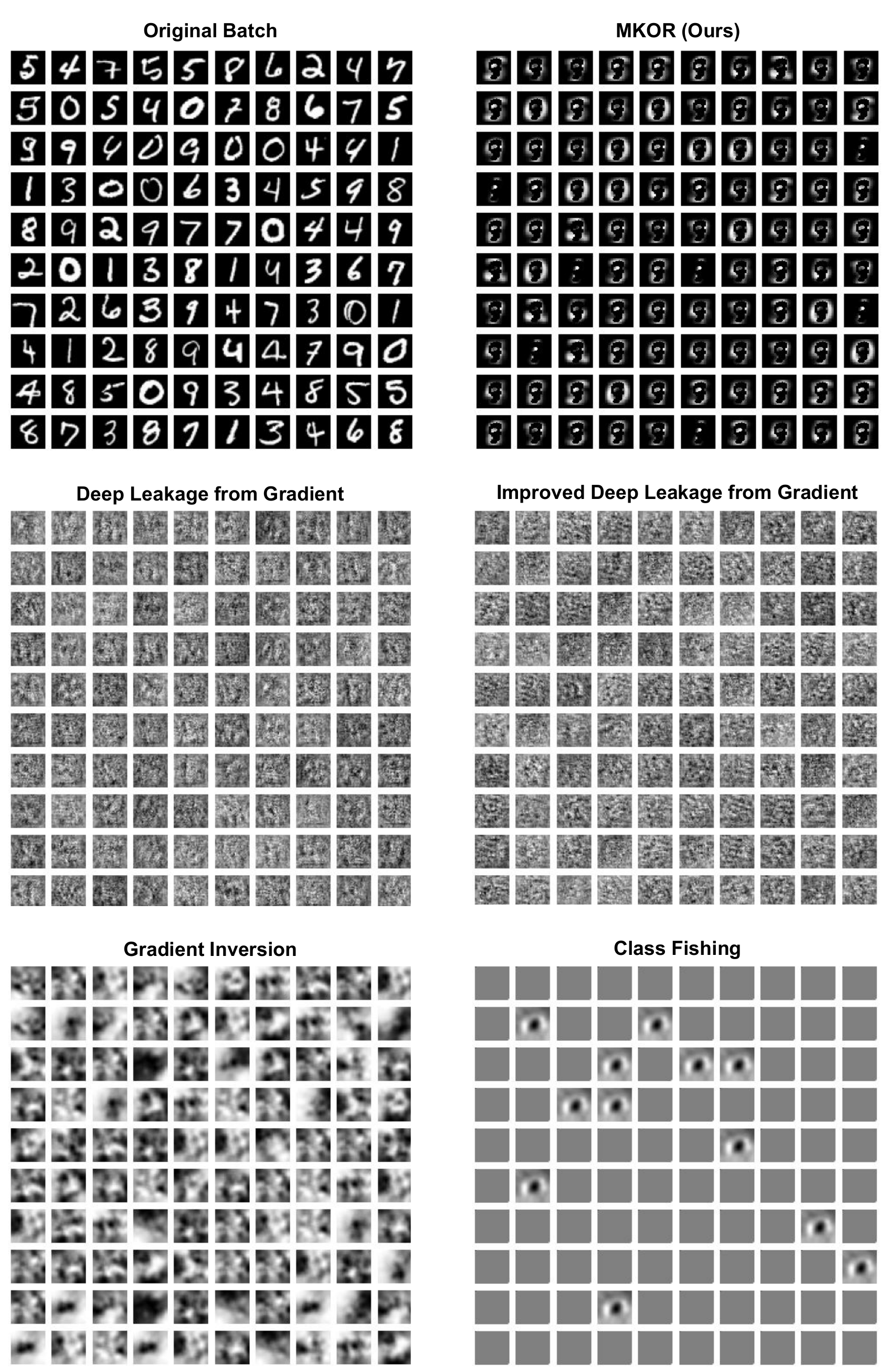}
 \vspace{-0.3cm}
	\caption{Qualitative comparison between different input reconstruction methods under Soteria defense on MNIST with original LeNet5 for a batch size of 100.
 }
	\label{fig:MNIST_random100_originalLenet_Soteria}
\end{figure*}

To show the robustness and effectiveness of our MKOR attack, we test MKOR along with other attack methods against two state-of-the-art defensive strategies. In Fig.~\ref{fig:MNIST_random100_originalLenet_Gaussian} and Fig.~\ref{fig:MNIST_random100_originalLenet_Soteria}, we show the comparison with original LeNet5 under Gaussian noise with normalized standard deviation $10^{-2}$ and Soteria defense with 30\% prune rate, respectively. The average numerical  results are shown in Table \ref{tab:Gaussian_random_bs100} and Table \ref{tab:Soteria_random_bs100}. 
The gradient clipping only changes the average of our reconstruction and does not affect the MKOR reconstruction result, while the additive artificial noise needs to be so high that it diminishes the model accuracy if the server is benign. 
Furthermore, Soteria and other optimization-based defensive detection schemes take longer time than gradient generation itself, thus can demand high edge device memory, computation and energy resources, and might potentially flag natural examples as well.
We believe that a comprehensive defense approach by measuring the level of parameter inconspicuousness, such as sparse representation and low-rank representation, can be developed as more attack methods are proposed.

\begin{table}[t!]
    \small
	\begin{center}
	    \caption{\small{Comparison with defensive Gaussian noise on gradients on MNIST with LeNet5 for random batches with batch size 100.}}
     \resizebox{0.85\linewidth}{!}{%
		\begin{tabular}{ | l |  l | l | } \hline
			\vtop{\hbox{\strut \emph{max}, avg SSIM}\hbox{\strut \emph{max}, avg PSNR}} & \vtop{\hbox{\strut Original LeNet}\hbox{\strut Gaussian $(10^{-2})$}} & \vtop{\hbox{\strut Modified LeNet}\hbox{\strut Gaussian $(10^{-1})$}} \\   \hline
            DLG & \emph{0.10}, 0.01, \emph{-3.53}, -5.58 & \emph{0.01}, 0.00, \emph{-6.42}, -11.89 \\   \hline
            iDLG & \emph{0.14}, 0.04, \emph{0.59}, -0.59 & \emph{0.01}, 0.00, \emph{-2.60}, -14.46 \\   \hline
            GI & \emph{0.29}, 0.12, \emph{15.44}, 11.23 & \emph{0.28}, 0.07, \emph{9.88}, 6.08 \\   \hline
            CF & \emph{0.20}, —, \emph{14.42}, — & \emph{0.33}, —, \emph{12.52}, — \\   \hline
            MKOR (ours) & \textbf{0.46}, 0.25, \textbf{16.71}, 12.74 & \textbf{0.63}, 0.37, \textbf{18.25}, 13.41 \\   \hline
		\end{tabular}
  }
		\label{tab:Gaussian_random_bs100}
	\end{center}
 \vspace{-0.4cm}
\end{table}

\begin{table}[b!]
    \small
	\begin{center}
	    \caption{Comparison with Soteria defense on MNIST with LeNet5 for random batches with batch size 100.}
     \resizebox{1\linewidth}{!}{%
		\begin{tabular}{ | l |  l | l | } \hline
			\vtop{\hbox{\strut \emph{max}, avg SSIM}\hbox{\strut \emph{max}, avg PSNR}} & \vtop{\hbox{\strut Original LeNet}\hbox{\strut Prune rate $30\%$}} & \vtop{\hbox{\strut Modified LeNet}\hbox{\strut Prune rate $30\%$}} \\   \hline
            DLG & \emph{0.00}, 0.00, \emph{-32.15}, -35.14 & \emph{0.01}, 0.00, \emph{-13.41}, -24.15 \\   \hline
            iDLG & \emph{0.00}, 0.00, \emph{-30.75}, -33.49 & \emph{0.00}, 0.00, \emph{-18.07}, -25.66 \\   \hline
            GI & \emph{0.19}, 0.01, \emph{7.89}, 2.77 & \emph{0.09}, 0.01, \emph{2.89}, -1.81 \\   \hline
            CF & \emph{0.16}, —, \emph{11.43}, — & \emph{0.17}, —, \emph{8.73}, — \\   \hline
            MKOR (ours) & \textbf{0.34}, 0.12, \textbf{15.29}, 10.66 & \textbf{0.71}, 0.38, \textbf{18.79}, 13.49 \\   \hline
		\end{tabular}
  }
		\label{tab:Soteria_random_bs100}
	\end{center}
 \vspace{-0.4cm}
\end{table}

In Fig.~\ref{fig:outputs_unique_1}, Fig.~\ref{fig:outputs_unique_2}, and Fig.~\ref{fig:outputs_unique_4}, we present three other unique batches of CIFAR-100 reconstructed by MKOR via VGG16, where each image is the only one representing its class, and we compare with the original batches and those reconstructed by GI, and CF. As can be seen, overall, the images reconstructed by GI are much more distorted compared to proposed MKOR. Moreover, CF can only reconstruct one image per batch without necessarily providing the best reconstruction performance on this one image. Furthermore, as shown in Table 3 of our main paper,  the image in the batch reconstructed by MKOR with the best performance has higher SSIM and PSNR values than the only image reconstructed by class fishing.

\begin{figure*}[h]
	\centering
	\includegraphics[width=0.7\linewidth]{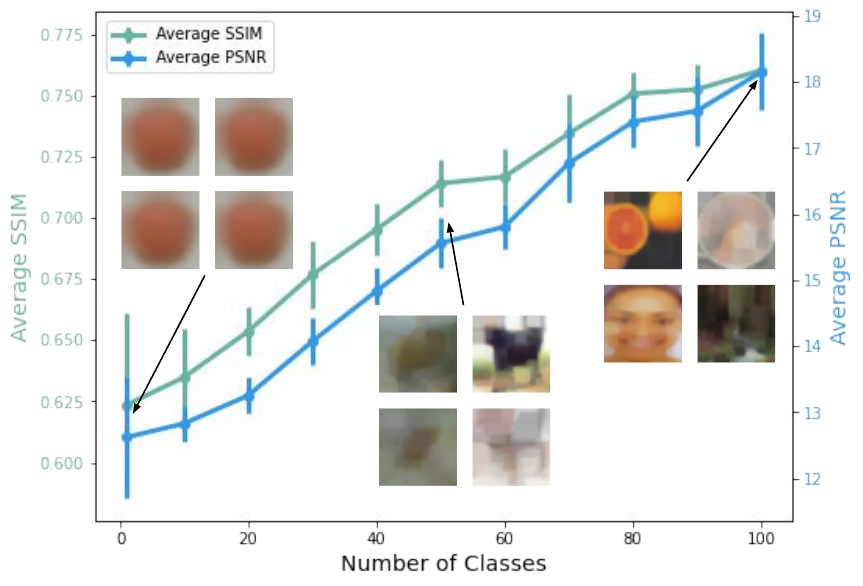}
 \vspace{-0.1cm}
	\caption{Reconstruction performance of MKOR on batches with at most $N$ classes. Each data point shows the mean and standard deviation over $U=10$ batches. We also plot 4 out of all $K=100$ samples in a batch for the cases of $N=1$, $N=50$ and $N=100$.
 }
	\label{fig:performance_vs_unique_class}
  \vspace{-0.2cm}
\end{figure*}

\begin{figure*}[h]
  \centering
  \includegraphics[width=1\textwidth]{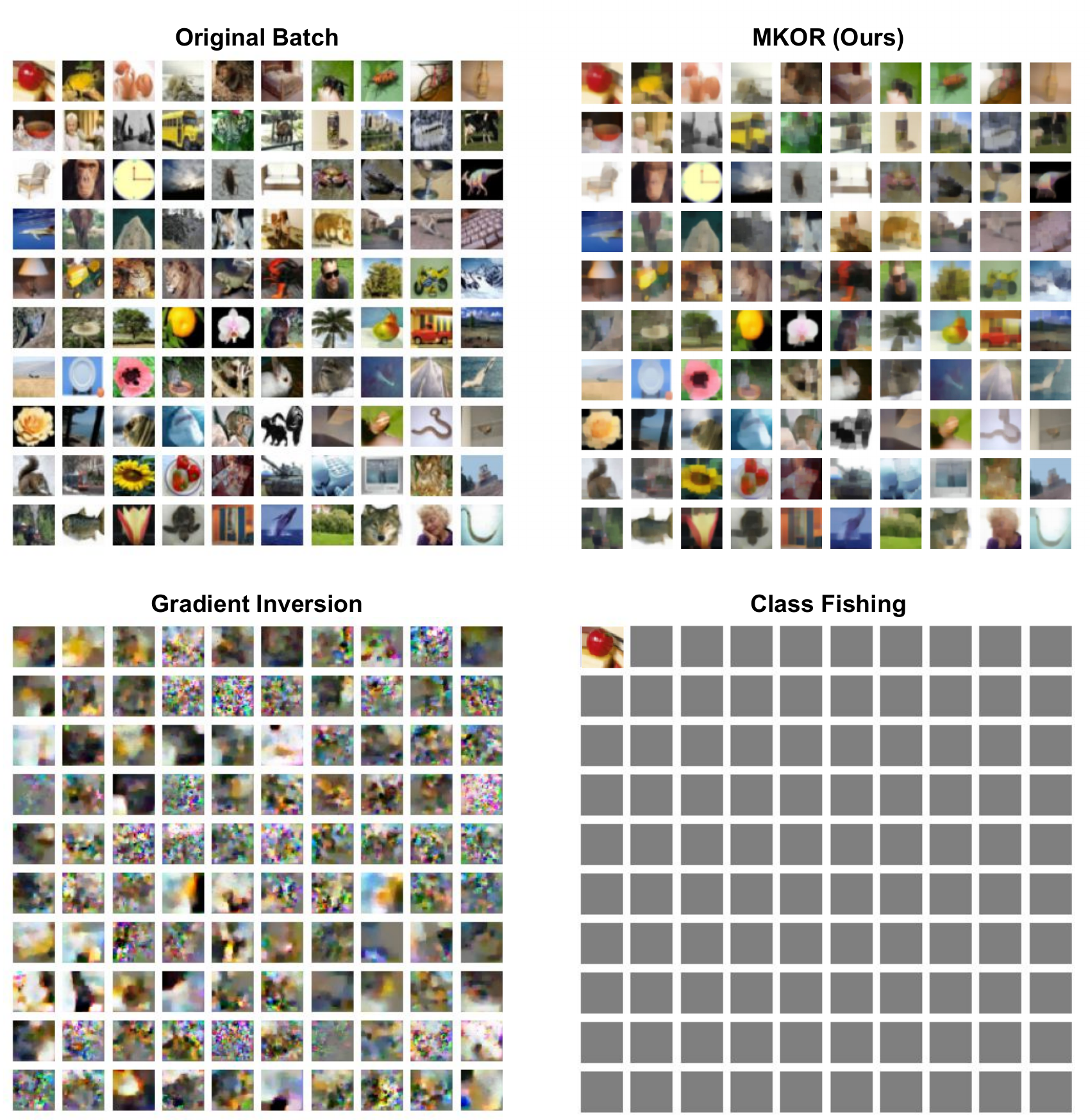}
  \caption{Qualitative comparison between different input reconstruction methods on CIFAR-100 for an additional batch with a batch size of 100.}
  \label{fig:outputs_unique_1}
\end{figure*}

\begin{figure*}[h]
  \centering
  \includegraphics[width=1\textwidth]{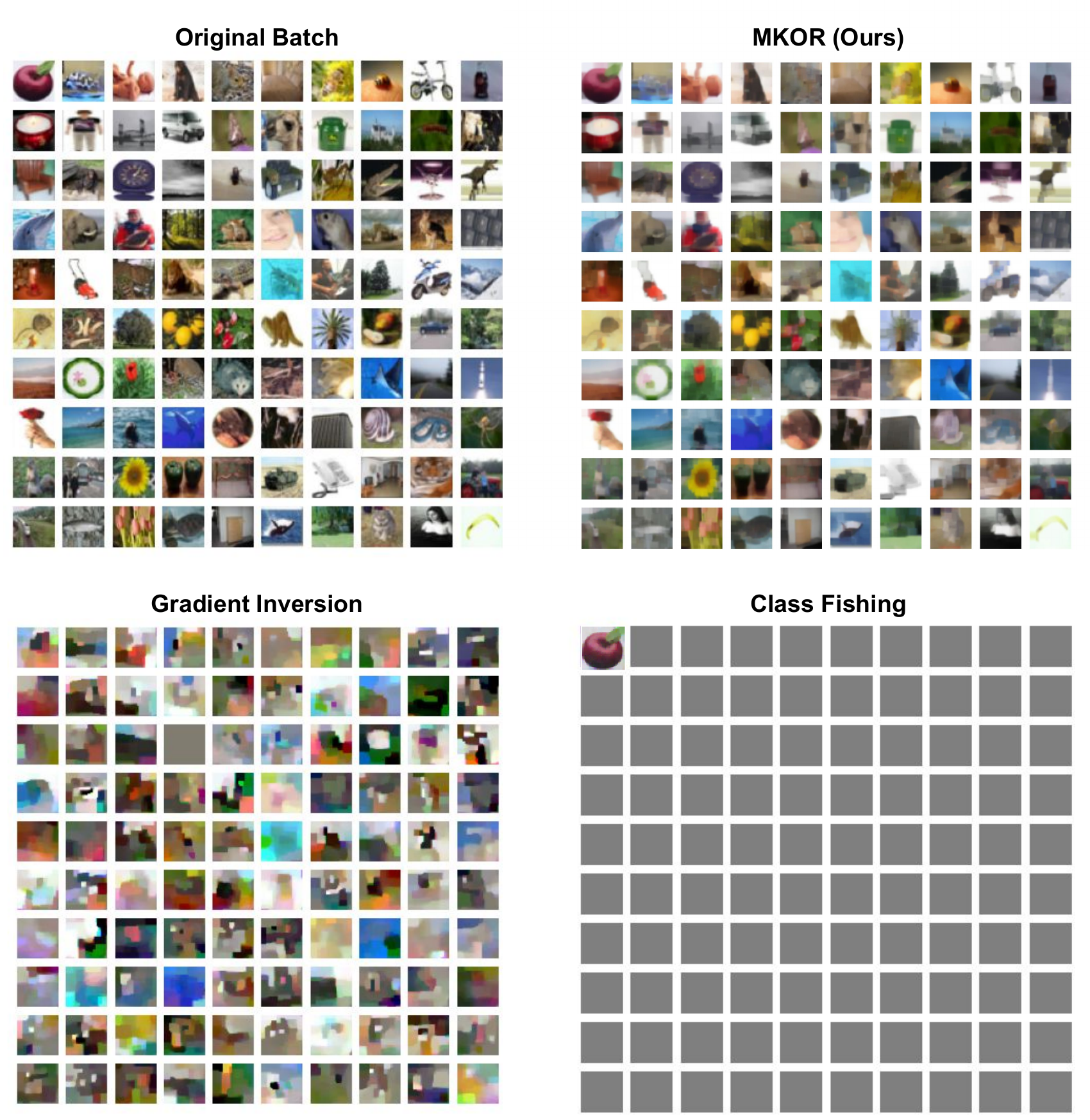}
  \caption{Qualitative comparison between different input reconstruction methods on CIFAR-100 for another additional batch with a batch size of 100.}
  \label{fig:outputs_unique_2}
\end{figure*}

\begin{figure*}[h]
  \centering
  \includegraphics[width=1\textwidth]{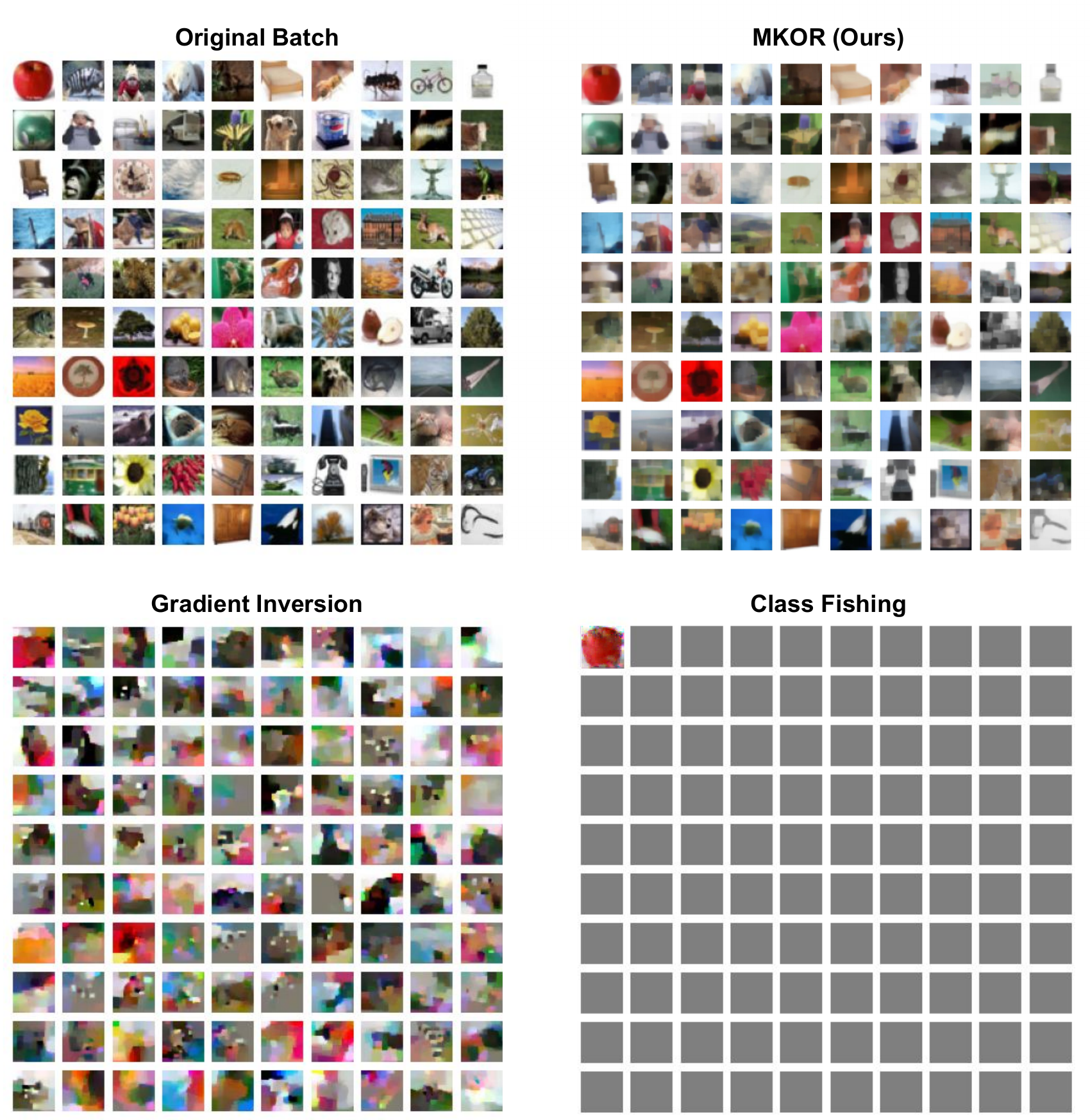}
  \caption{Qualitative comparison between different input reconstruction methods on CIFAR-100 for another additional batch with a batch size of 100.}
  \label{fig:outputs_unique_4}
\end{figure*}

While these approaches achieve the above results with unique CIFAR-100 samples, their performance can degrade when there are multiple images from the same class. CF is especially vulnerable, since it can only reconstruct one image per batch. In~Fig.~\ref{fig:performance_vs_unique_class}, performance of MKOR is shown when samples are randomly selected from at most $N$ classes, allowing different images from the same class. The number of unique samples (only ones from their class) is estimated by Eq.~(3) in the paper.
The performance degrades as $N$ decreases, e.g., when $N\!\!=\!\!1$, reconstruction is the average of $K=100$ images belonging to the same class. We note that the performance variance between batches for the same $N$ value is due to the difference in data but not the MKOR algorithm.

\begin{figure*}[h]
        \centering
	\includegraphics[width=0.75\linewidth]{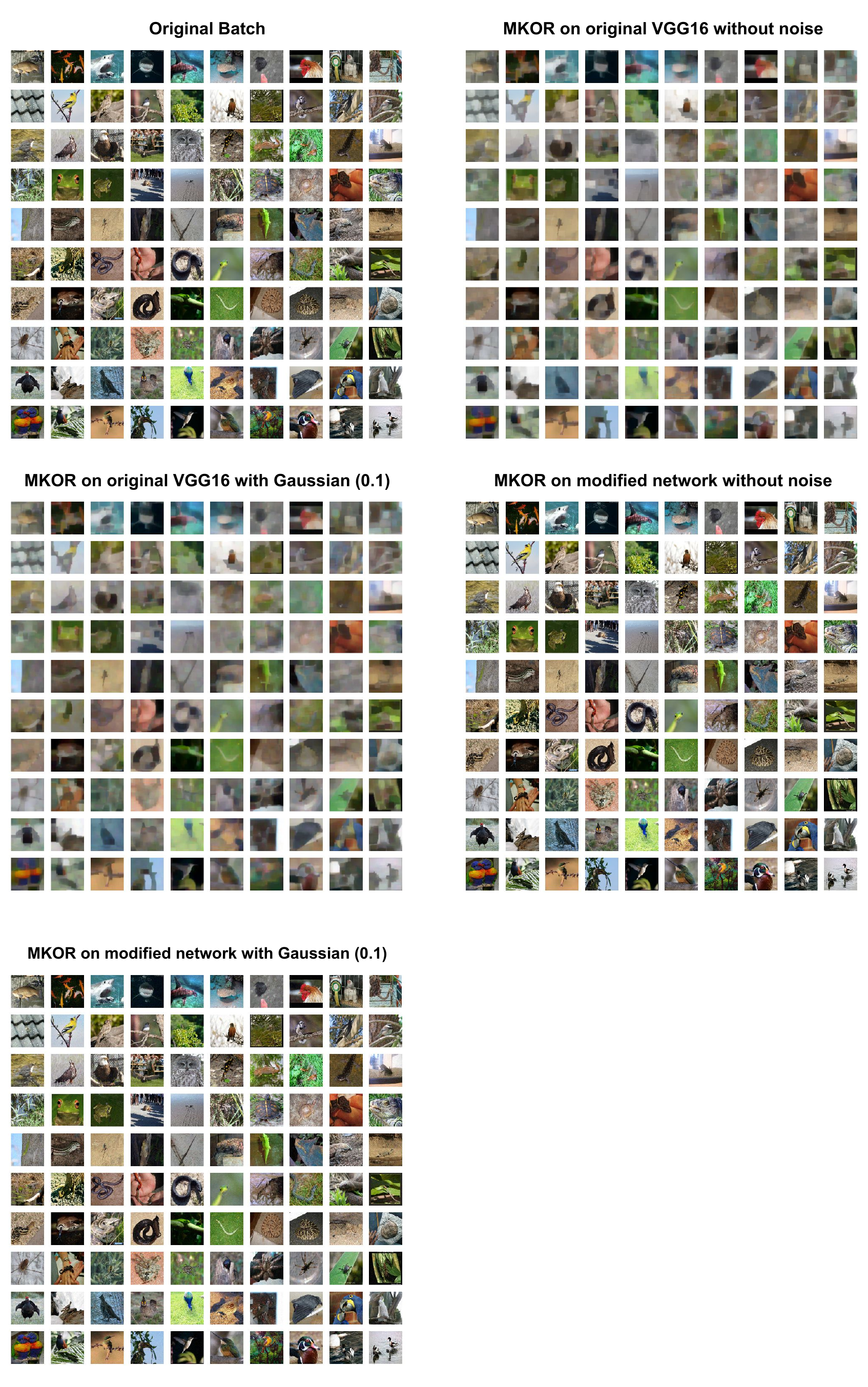}
 \vspace{-0.3cm}
	\caption{Input reconstruction on a \textbf{unique} batch with 1000 ImageNet images, with either original VGG16 or modified network, and with either no noise or $10^{-1}$ Gaussian noise. We randomly display 100 images out of 1000 in the batch for simplicity.
 }
	\label{fig:ImageNet_unique1000}
\end{figure*}

\begin{figure*}[h]
        \centering
	\includegraphics[width=0.75\linewidth]{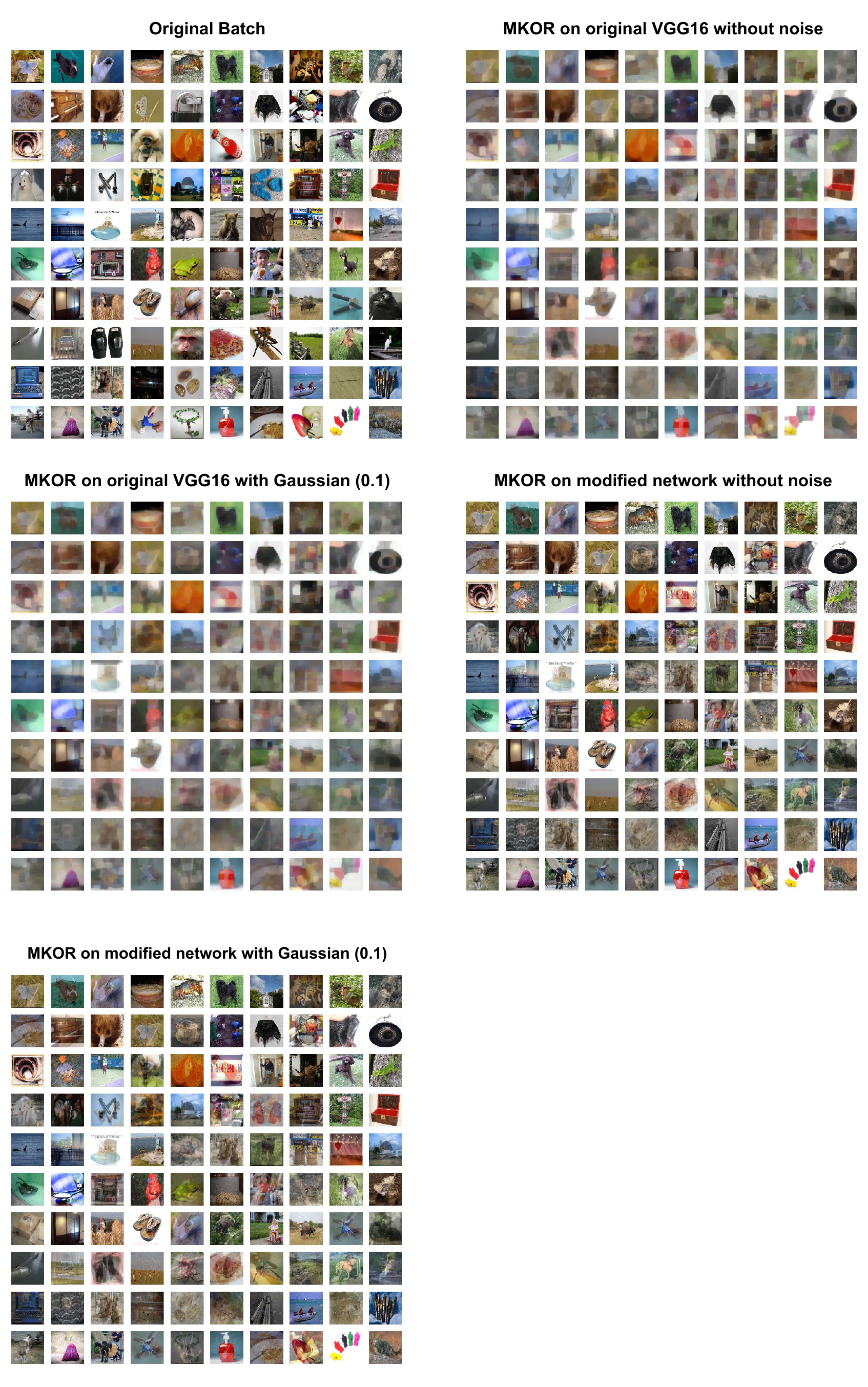}
	\caption{Input reconstruction on a \textbf{random} batch with 1000 ImageNet images, with either original VGG16 or modified network, and with either no noise or $10^{-1}$ Gaussian noise. We randomly display 100 images out of 1000 in the batch for simplicity.
 }
	\label{fig:ImageNet_random1000}
\end{figure*}

In Fig.~\ref{fig:ImageNet_unique1000}, we present the MKOR reconstruction on a unique batch with 1000 ImageNet images with original VGG16 network and modified network, respectively. In Fig.~\ref{fig:ImageNet_random1000}, we present the reconstruction on a random batch with the same setting. We notice that the batch is very large while the high-resolution image has many details, so we randomly select 100 images in the batch to show in the plots.

\begin{figure*}[h]
	\centering
	\includegraphics[width=1\linewidth]{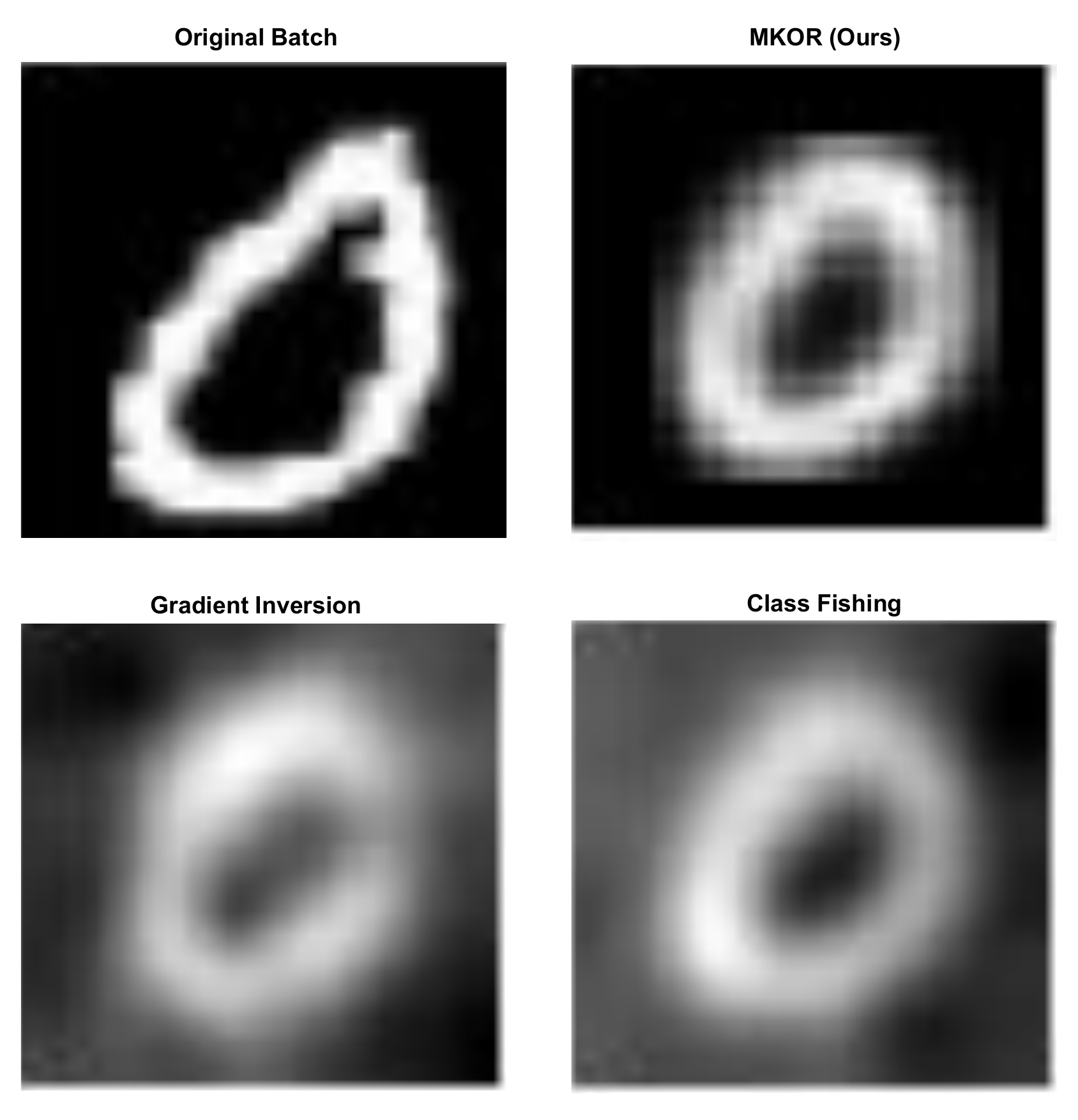}
	\caption{Single samples in batches of different input reconstruction methods on MNIST with original LeNet5 for a random batch of size 100. }
 \label{fig:SingleSample_MNIST_random100_originalLenet_only4}
\end{figure*}

\begin{figure*}[h]
	\centering
	\includegraphics[width=0.8\linewidth]{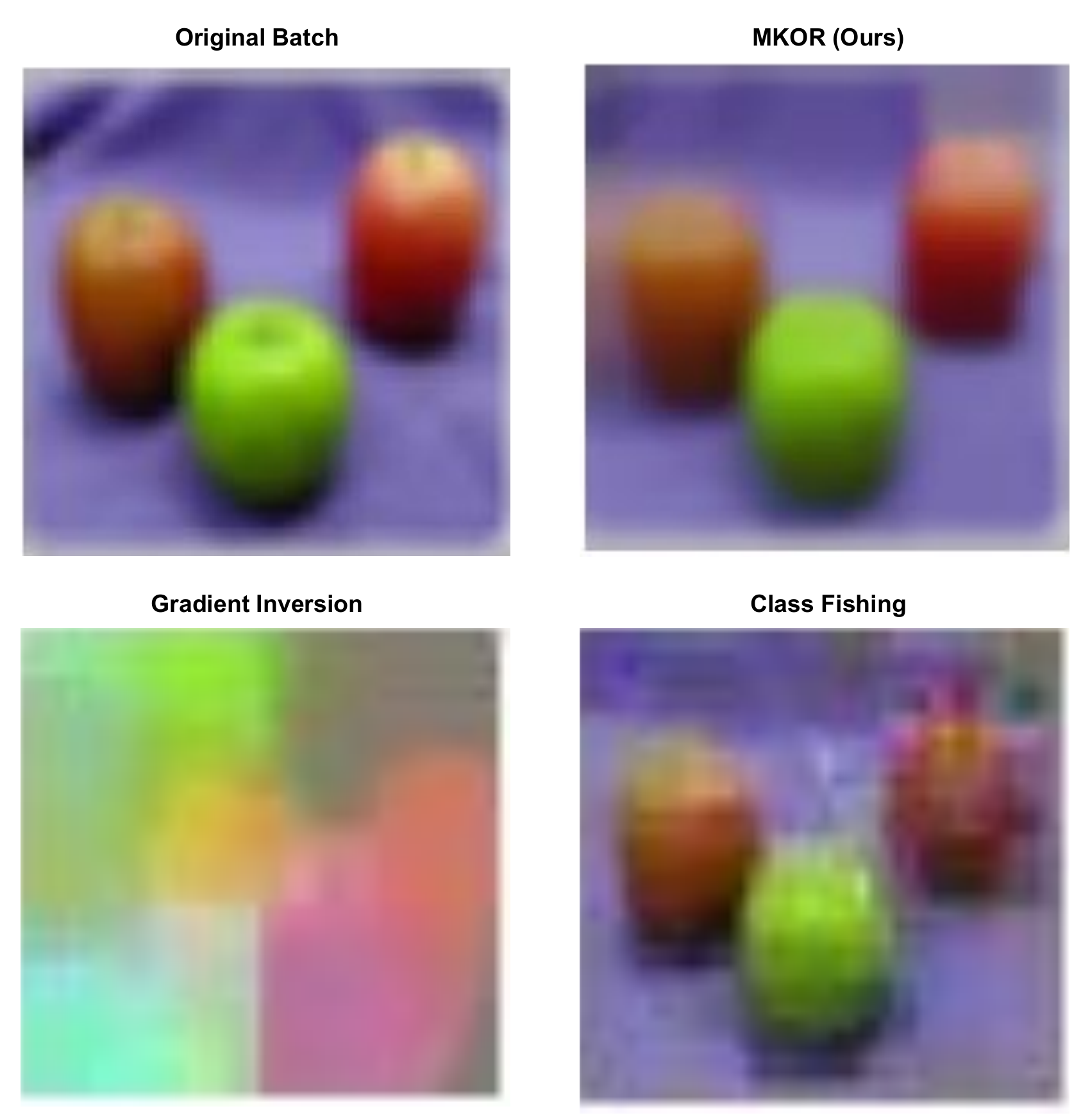}
 \vspace{-0.3cm}
	\caption{Single samples in batches of different input reconstruction methods on CIFAR-100 for a unique batch of size 100.
 }
	\label{fig:SingleSample_outputs_copy_gradinver_input}
\end{figure*}

\begin{figure*}[h]
        \centering
	\includegraphics[width=0.75\linewidth]{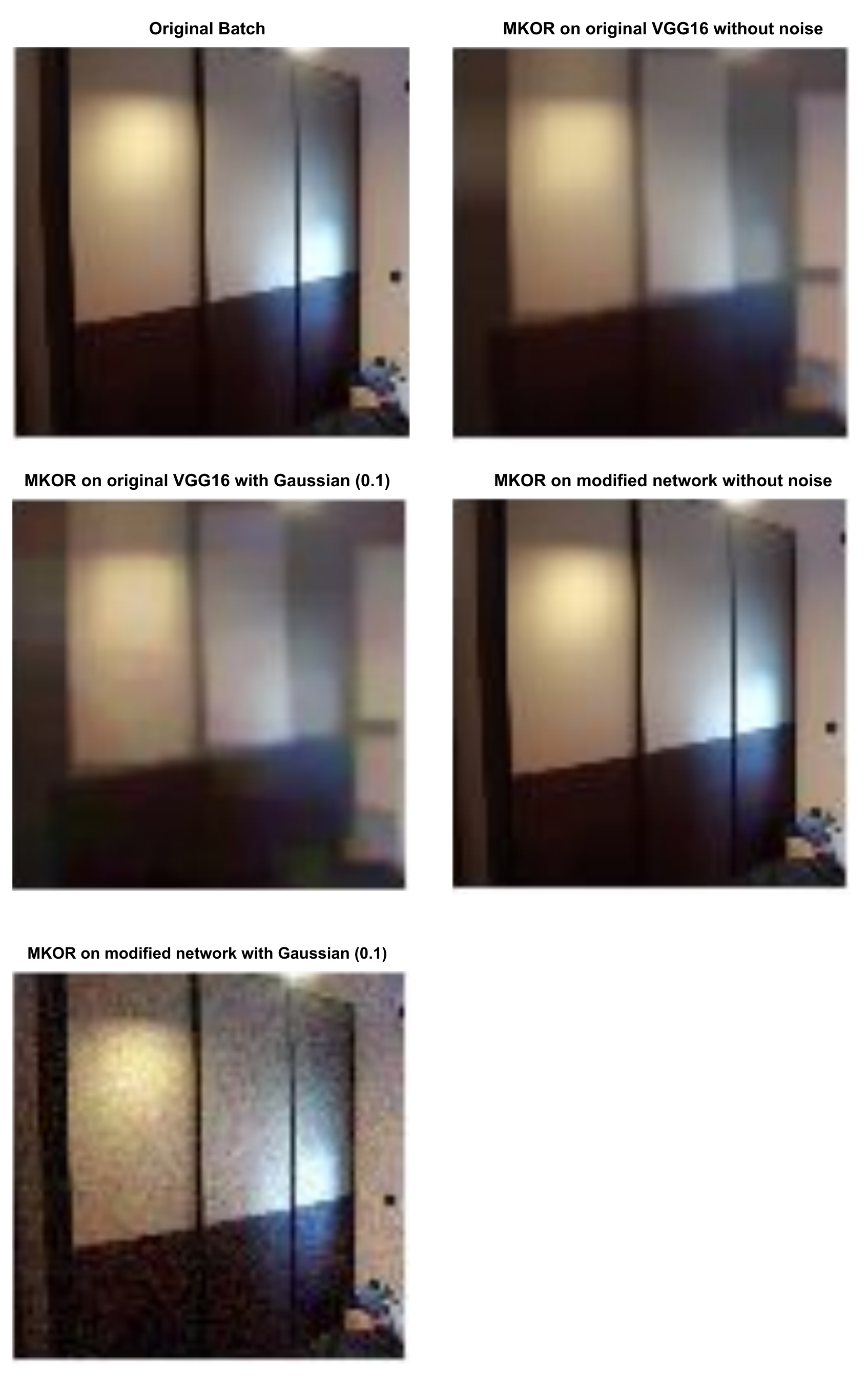}
 \vspace{-0.3cm}
	\caption{Single samples in random batches with 1000 ImageNet images, with either original VGG16 or modified network, and with either no noise or $10^{-1}$ Gaussian noise. .
 }
	\label{fig:SingleSample_ImageNet_random1000}
\end{figure*}

To show the reconstruction details, we present larger single samples in Figs.~\ref{fig:SingleSample_MNIST_random100_originalLenet_only4}, \ref{fig:SingleSample_outputs_copy_gradinver_input}, \ref{fig:SingleSample_ImageNet_random1000}, respectively, for Fig. 3 in the main paper, Fig. 4 in the main paper, and Fig.~\ref{fig:ImageNet_random1000}.

We also provide our code as a zipped supplementary file. The implementation details of the proposed MKOR can be found in this zipped file, the implementation details of DLG and iDLG can be found in \cite{zhao2020idlg}, and the implementation details of GI and CF can be found in \cite{Jonas2023}.


\begin{multicols}{2}
\end{multicols}
\begin{multicols}{2}
\end{multicols}

\begin{multicols}{2}
\end{multicols}
\begin{multicols}{2}
\end{multicols}

\begin{multicols}{2}
\end{multicols}
\begin{multicols}{2}
\end{multicols}

\begin{multicols}{2}
\end{multicols}
\begin{multicols}{2}
\end{multicols}

\begin{multicols}{2}
\end{multicols}
\begin{multicols}{2}
\end{multicols}

\vspace{7cm}
\begin{multicols}{2}
\end{multicols}
\begin{multicols}{2}
\end{multicols}

\begin{multicols}{2}
\end{multicols}
\begin{multicols}{2}
\end{multicols}

\begin{multicols}{2}
\end{multicols}
\begin{multicols}{2}
\end{multicols}




\begin{multicols}{2}
\end{multicols}
\begin{multicols}{2}
\end{multicols}

\begin{multicols}{2}
\end{multicols}
\begin{multicols}{2}
\end{multicols}

\begin{multicols}{2}
\end{multicols}
\begin{multicols}{2}
\end{multicols}

\begin{multicols}{2}
\end{multicols}
\begin{multicols}{2}
\end{multicols}

\begin{multicols}{2}
\end{multicols}
\begin{multicols}{2}
\end{multicols}

\end{document}